%% file: main.tex
\documentclass[lettersize,journal]{IEEEtran}
\usepackage{amsmath,amsfonts}
\usepackage{algorithmic}
\usepackage{algorithm}
\usepackage{array}
\usepackage[caption=false,font=normalsize,labelfont=sf,textfont=sf]{subfig}
\usepackage{textcomp}
\usepackage{stfloats}
\usepackage{url}
\usepackage{verbatim}
\usepackage{graphicx}
\usepackage{cite}
\usepackage{tabulary}
\usepackage[dvipsnames]{xcolor}
\usepackage{booktabs}
\usepackage{multirow}
\usepackage[breaklinks,colorlinks]{hyperref}
\newcommand{\tablestyle}[2]{\setlength{\tabcolsep}{#1}\renewcommand{\arraystretch}{#2}\centering\footnotesize}
\begin{document}

\title{Oriented Tiny Object Detection: A Dataset, Benchmark, and Dynamic Unbiased Learning}

\author{Chang Xu$^{*}$, Ruixiang Zhang$^{*}$, Wen Yang$^{\dag}$, Haoran Zhu, Fang Xu, Jian Ding, Gui-Song Xia

\thanks{* Equal contribution, $^{\dag}$ Corresponding author.}
\thanks{C. Xu, R. Zhang, W. Yang, H. Zhu, and F. Xu are with the School of Electronic Information, Wuhan University, Wuhan, 430072 China. \emph{E-mail: \{xuchangeis, zhangruixiang, yangwen, zhuhaoran, xufang\}@whu.edu.cn}}
\thanks{J, Ding is with the King Abdullah University of Science and Technology. \emph{E-mail: jian.ding@kaust.edu.sa}}
\thanks{G-S. Xia is with the School of Computer Science and the State Key Lab. LIESMARS, Wuhan University, Wuhan, 430072, China. \emph{E-mail: guisong.xia@whu.edu.cn}}
}

\markboth{Journal of \LaTeX\ Class Files,~Vol.~14, No.~8, August~2021}%
{Shell \MakeLowercase{\textit{et al.}}: A Sample Article Using IEEEtran.cls for IEEE Journals}

\maketitle

\begin{abstract}
Detecting oriented tiny objects, which are limited in appearance information yet prevalent in real-world applications, remains an intricate and under-explored problem. 
To address this, we systemically introduce a new dataset, benchmark, and a dynamic coarse-to-fine learning scheme in this study. 
Our proposed dataset, AI-TOD-R, features the smallest object sizes among all oriented object detection datasets. Based on AI-TOD-R, we present a benchmark spanning a broad range of detection paradigms, including both fully-supervised and label-efficient approaches. Through investigation, we identify a learning bias presents across various learning pipelines: confident objects become increasingly confident, while vulnerable oriented tiny objects are further marginalized, hindering their detection performance.
To mitigate this issue, we propose a Dynamic Coarse-to-Fine Learning (DCFL) scheme to achieve unbiased learning.
DCFL dynamically updates prior positions to better align with the limited areas of oriented tiny objects, and it assigns samples in a way that balances both quantity and quality across different object shapes, thus mitigating biases in prior settings and sample selection.
Extensive experiments across eight challenging object detection datasets demonstrate that DCFL achieves state-of-the-art accuracy, high efficiency, and remarkable versatility. The dataset, benchmark, and code are available at \url{https://chasel-tsui.github.io/AI-TOD-R/}.
\end{abstract}


\input{introduction}

\input{relatedworks}
\input{aitodr}
\input{benchmark}
\input{method}
\input{experiments}

\section{Conclusion}
In this work, we systematically address the challenging task of detecting oriented tiny objects by establishing a new dataset and a benchmark, and proposing a dynamic coarse-to-fine learning scheme aimed at scale-unbiased learning. Our dataset, AI-TOD-R, has the smallest mean object size among all oriented object detection datasets, and it presents additional challenges such as dense arrangement and class imbalance. Based on this dataset, we establish a benchmark and investigate the performance of various detection paradigms, uncovering two key insights.
First, label-efficient detection methods now offer highly competitive performance on oriented tiny objects, showing great potential for further exploration. Second, biased prior settings and biased sample assignment across various detection pipelines significantly impede the detection performance of oriented tiny objects. To address these biases, we propose a dynamic coarse-to-fine learning (DCFL) scheme that is applicable to both one-stage and two-stage architectures. Extensive experiments on eight heterogeneous benchmarks verify that DCFL can significantly improve the detection accuracy of oriented tiny objects while maintaining high efficiency.

\section*{Acknowledgments}
We would like to thank Zijuan Chen, Xianhang Ye, Nuoyi Wang, Jinrui Zhang, Yuxin Li, Zheyan Xiao, Ziming Gui, Zhiwei Chen, Zijun Wu, and Huan Li for their voluntary annotation works. This work was supported in part by the National Natural Science Foundation of China (NSFC) under Grant 62271355.

\bibliographystyle{IEEEtran}
\bibliography{chang,jinwang}

\newpage

\vspace{-33pt}
\begin{IEEEbiography}[{\includegraphics[width=1in,height=1.25in,clip,keepaspectratio]{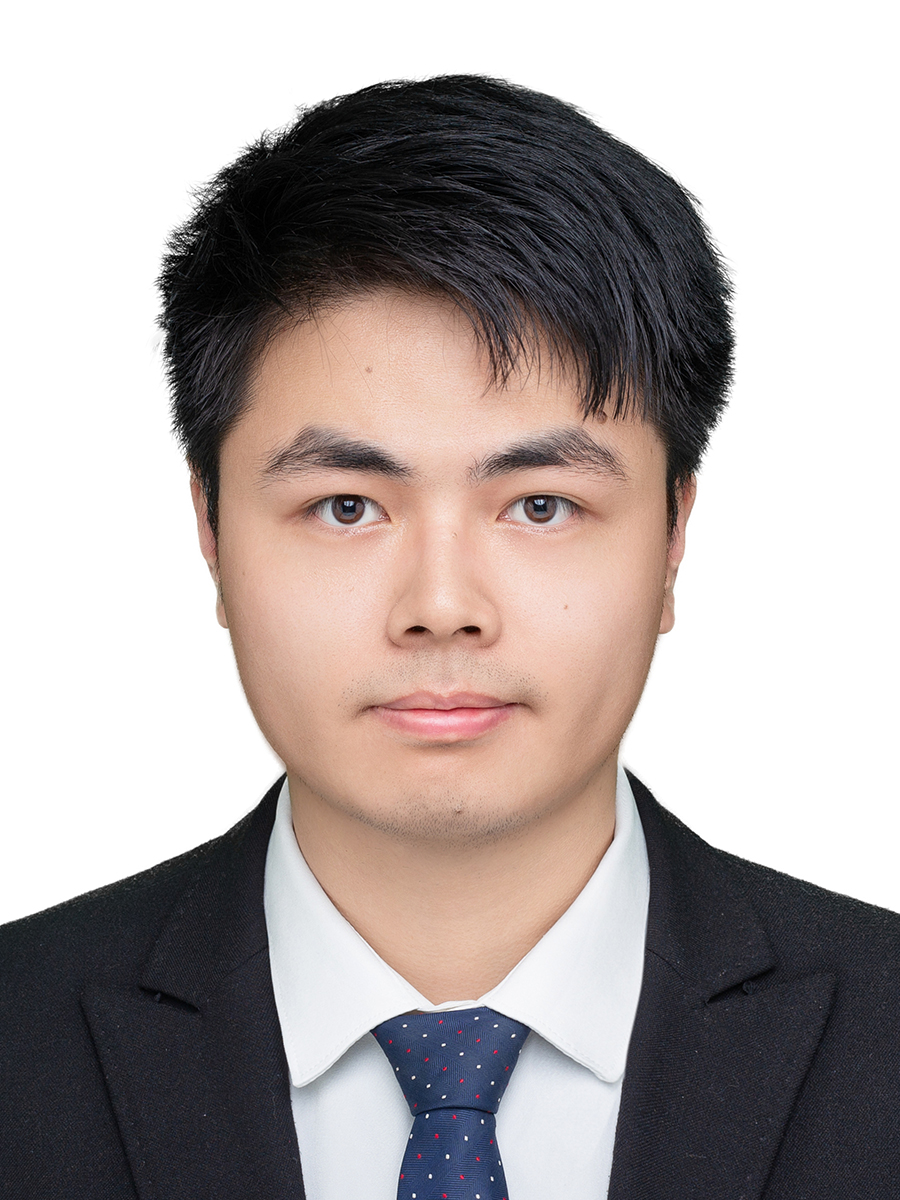}}]{Chang Xu}
received his B.S. degree in electronic information engineering, and his M.S. degree in information and communication systems, both from Wuhan University, Wuhan, China, in 2021, and 2024, respectively. This work was done during his master study in Wuhan University.
He is currently pursuing his Ph.D. degree in the Environmental Computational Science and Earth Observation Laboratory, EPFL, Sion, Switzerland. 
His research focuses on object detection, visual geo-localization, and multi-modal learning.
\end{IEEEbiography}
\begin{IEEEbiography}[{\includegraphics[width=1in,height=1.25in,clip,keepaspectratio]{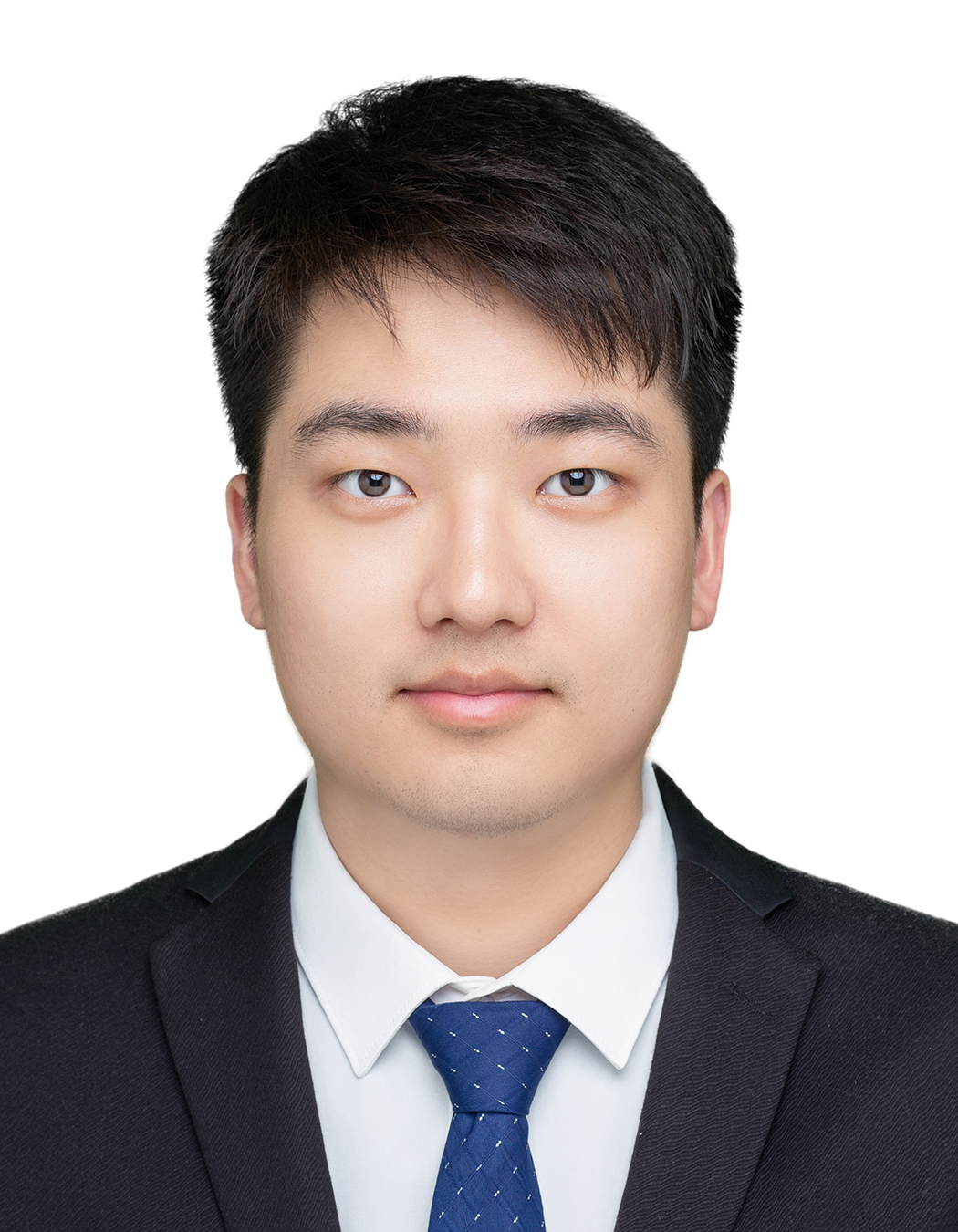}}]{Ruixiang Zhang}
received the B.S. in electronic engineering from Wuhan University, China, in 2019. He is currently working towards a Ph.D degree in communication and information system at Wuhan University, China. His research involves remote sensing image processing, including label-efficient object detection and cross-modal object detection.
\end{IEEEbiography}
\begin{IEEEbiography}[{\includegraphics[width=1in,height=1.25in,clip,keepaspectratio]{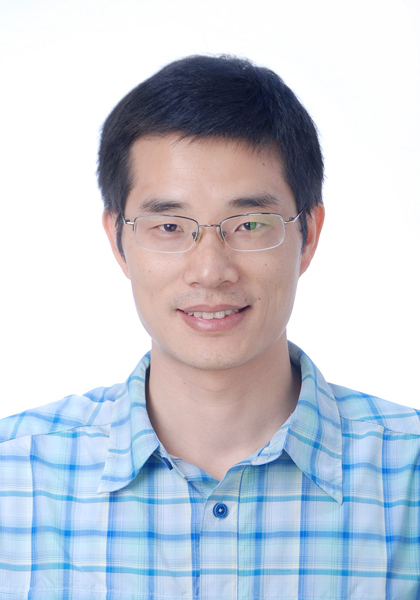}}]{Wen Yang}
(Senior Member, IEEE)~received his B.S. degree in Electronic Apparatus and Surveying Technology, his M.S. degree in Computer Application Technology, and his Ph.D. degree in Communication and Information System, all from Wuhan University, Wuhan, China, in 1998, 2001, and 2004, respectively. In 2008 and 2009, he worked as a Visiting Scholar with the Apprentissage et Interfaces (AI) Team at the Laboratoire Jean Kuntzmann in Grenoble, France. Following that, he served as a Post-Doctoral Researcher with the State Key Laboratory of Information Engineering, Surveying, Mapping, and Remote Sensing, also at Wuhan University, from 2010 to 2013. Since then, he has held the position of Full Professor at the School of Electronic Information, Wuhan University. He was also a guest professor of the Future Lab AI4EO at the Technical University of Munich (TUM). He received the U.V. Helava Award for the best paper in the {\em ISPRS Journal of Photogrammetry and Remote Sensing} in 2021. His research interests include object detection and recognition, multisensor information fusion, and remote sensing image interpretation.
\end{IEEEbiography}
\begin{IEEEbiography}[{\includegraphics[width=1in,height=1.25in,clip,keepaspectratio]{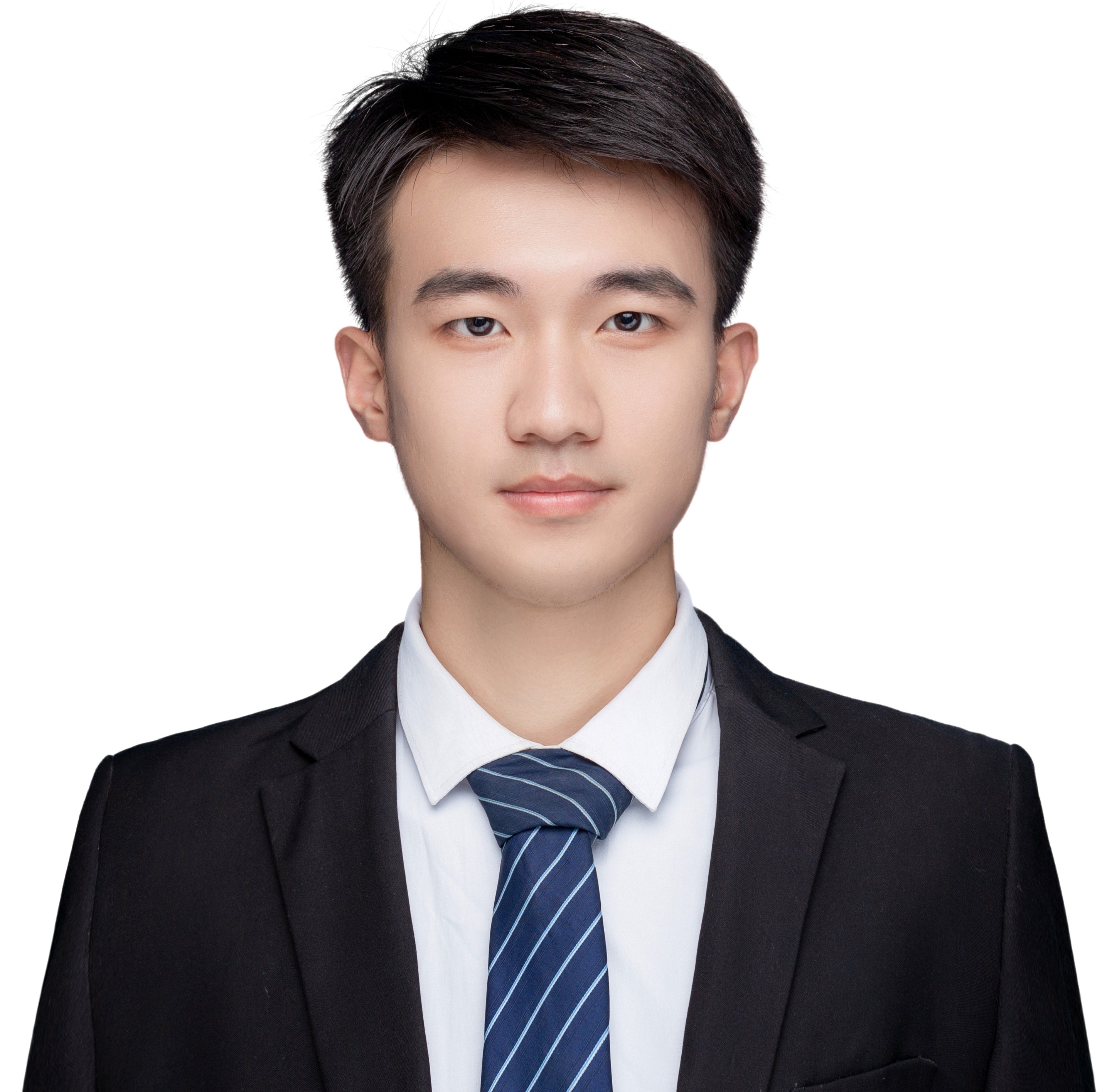}}]{Haoran Zhu}
received his B.S. degree in electronic information engineering from Wuhan University, Wuhan, China, in 2023, where he is currently pursuing the Ph.D. degree. His research interests include computer vision and remote sensing image tiny object detection.
\end{IEEEbiography}

\begin{IEEEbiography}[{\includegraphics[width=1in,height=1.25in,clip,keepaspectratio]{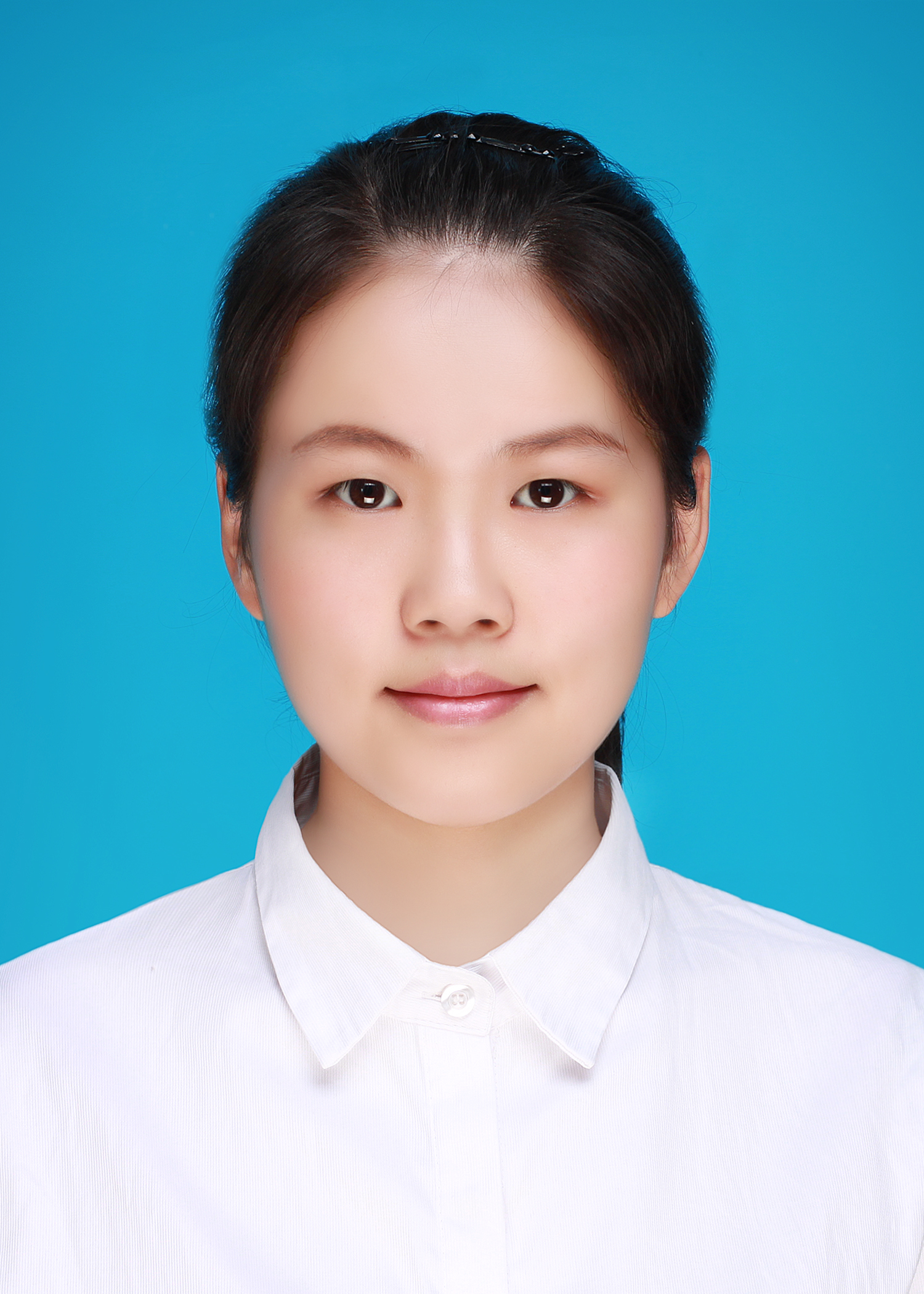}}]{Fang Xu}
received her B.S. degree in electronic and information engineering and her Ph.D. degree in communication and information system from Wuhan University, Wuhan, China, in 2018 and 2023, respectively. She is a postdoctoral researcher with the school of computer science, Wuhan University, China. Her research involves remote sensing image processing, including multi-modal data matching and fusion.
\end{IEEEbiography}

\begin{IEEEbiography}[{\includegraphics[width=1in,height=1.25in,clip,keepaspectratio]{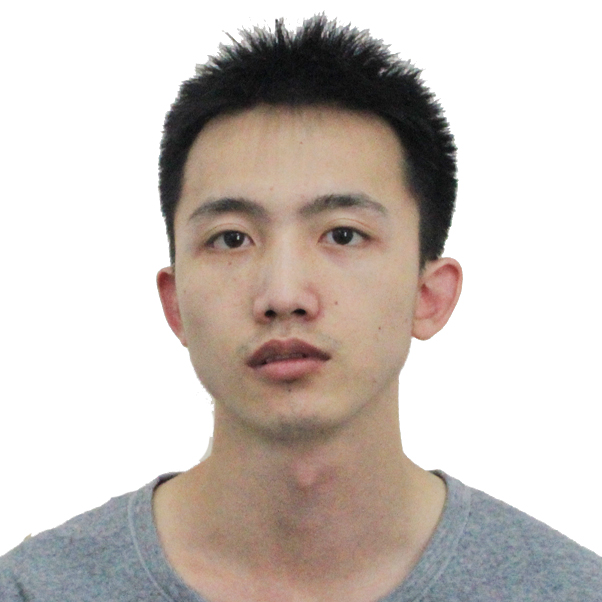}}]{Jian Ding} is currently a post-doctoral researcher in King Abdullah University of Science and Technology (KAUST). He received the B.S. degree in Aircraft Design and Engineering from Northwestern Polytechnical University, Xian, China in 2017, then obtained the Ph.D. degree at the State Key Laboratory of Information Engineering in Surveying, Mapping and Remote Sensing, Wuhan University, Wuhan, China in 2023. His research interests include object detection, instance segmentation, and remote sensing.
\end{IEEEbiography}

\begin{IEEEbiography}[{\includegraphics[width=1in,height=1.25in]{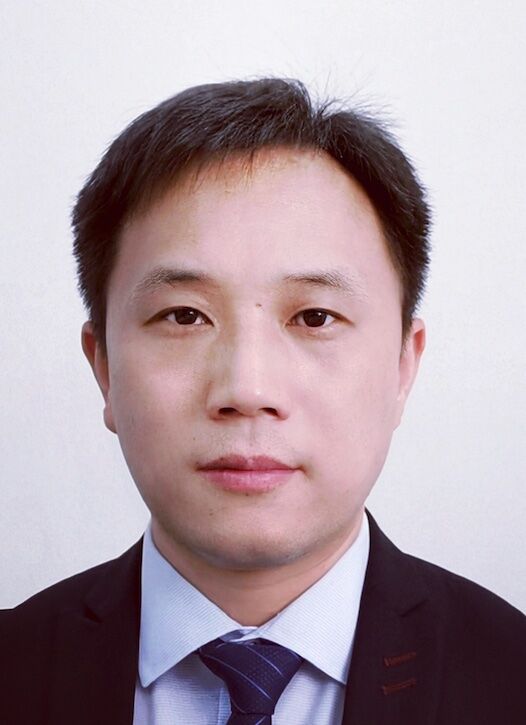}}]{Gui-Song Xia}
received his Ph.D. degree in image processing and computer vision from CNRS LTCI, T{\'e}l{\'e}com ParisTech, Paris, France, in 2011. From 2011 to 2012, he was a Post-Doctoral Researcher with the Centre de Recherche en Math{\'e}matiques de la Decision, CNRS, Paris-Dauphine University, Paris, for one and a half years.
He is currently working as a full professor in computer vision and photogrammetry at Wuhan University. He has also been working as a Visiting Scholar at DMA, {\'E}cole Normale Sup{\'e}rieure (ENS-Paris) for two months in 2018. He was also a guest professor of the Future Lab AI4EO at the Technical University of Munich (TUM). His current research interests include mathematical modeling of images and videos, structure from motion, perceptual grouping, and remote sensing image understanding. He serves on the Editorial Boards of several journals, including {\em ISPRS Journal of Photogrammetry and Remote Sensing, Pattern Recognition, Signal Processing: Image Communications, EURASIP Journal on Image \& Video Processing, Journal of Remote Sensing, and Frontiers in Computer Science: Computer Vision}.
\end{IEEEbiography}

%\vspace{11pt}

\vfill

\end{document}

%% file: introduction.tex
\section{Introduction}
\label{sec:intro}

\IEEEPARstart{W}{hen} observations approach the physical limit of a camera's properties (\textit{e.g.}, focal length and resolution), the captured images will inevitably contain objects at extremely tiny scales. This situation, though extreme, is prevalent in real-world applications ranging from micro-vision (\textit{e.g.}, medical and cell imaging~\cite{lesion_2019_miccai}) to macro-vision (\textit{e.g.}, drone and satellite imaging~\cite{visdrone_2021_pami,DOTA2.0_2021_pami}). 
In these professional domains, imaging typically adopts an overhead perspective to more accurately capture the primary features of the objects, resulting in them appearing in arbitrary orientations.

Detecting arbitrarily oriented tiny objects is a fundamental yet highly challenging step towards achieving an intelligent understanding of these scenarios. Meanwhile, numerous risk-sensitive applications demand the precise and robust detection of tiny objects with orientation information, to name a few, traffic monitoring~\cite{vehicle_pami_2023}, border surveillance~\cite{border_2019}, medical diagnostics~\cite{lesion_2019_miccai}, and defect identification~\cite{small_defect_2022_tim}. 
Unfortunately, previous studies mainly focus on detecting generic objects~\cite{object_detection_survey_2023_ieee} or arbitrarily oriented objects~\cite{rs_object_detection_2020_isprs}. When it comes to the more challenging task of detecting oriented tiny objects, existing methods often struggle to deliver satisfactory performance.
Typically, 77\% objects in DOTA-v2\cite{DOTA2.0_2021_pami} are in the size range of $10^{2}$-$50^{2}$ pixels, while the State-Of-The-Art (SOTA) performance~\cite{orientedrcnn_2021_iccv} is still lower than 30\% $\rm{AP^{@50:5:95}}$. As a hard nut to crack in the community, what makes things worse is the \textit{lack of task-specific datasets and benchmarks} designed to prompt the development of detection methods.
So far, there is only one recently released dataset (SODA-A)~\cite{soda_2023_pami} tailored to relevant study, while its research focus mainly lies on small-scale rather than tiny-scale objects\footnote{According to existing literature~\cite{COCO_2014_ECCV,AI-TOD_2020_ICPR}, small and tiny objects are defined as objects smaller than $32^{2}$ and $16^{2}$ pixels, respectively}. 

\begin{figure}[t]
\centering
\includegraphics[width=\linewidth]{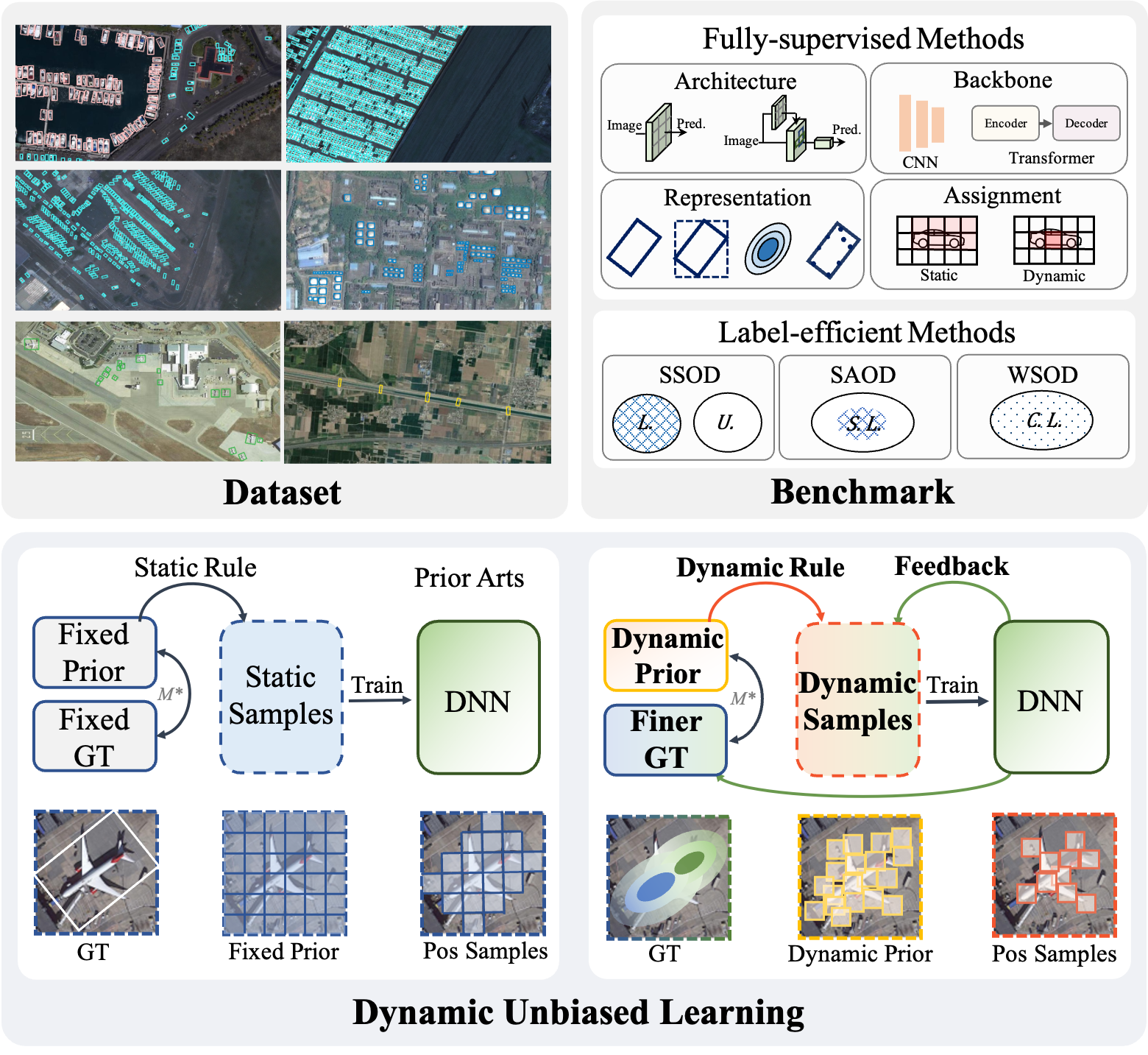}
\caption{This paper systemically introduces the challenging task of oriented tiny object detection, with the AI-TOD-R dataset, benchmark, and a dynamic coarse-to-fine learning pipeline. Upper: Typical annotation examples from AI-TOD-R and detection paradigms covered by this benchmark, where ``L.'', ``U.'', ``S. L.'', and ``C. L.'' denote labelled, unlabelled, sparsely labelled, and coarsely labelled images, respectively. Lower: A comparison of learning paradigms for oriented object detection. Compared to prior arts (left), our proposed pipeline (right) mitigates the learning bias against oriented tiny objects with a dynamically updated prior and a coarse-to-fine sample learning scheme.}
\label{fig1}
\end{figure}

% problems
Taking a step towards more severe challenges, this work systemically explores the oriented tiny object detection problem from perspectives of dataset, benchmark, and method. A visual summary of this work is presented in Figure~\ref{fig1}. To further push the boundaries of oriented object detection for extremely tiny objects, \textbf{we contribute a new dataset dedicated to oriented tiny object detection}, named AI-TOD-R (Section~\ref{sec:dataset}). With a mean object size of only 10.6$^{2}$ pixels, AI-TOD-R is the dataset of the smallest object size for oriented object detection. This challenging dataset is established using a semi-automatic annotation process via supplementing AI-TOD-v2~\cite{aitodv2_2022_isprs} with orientation information and ensuring high annotation quality. 
%AI-TOD-R solves the missing annotation, lack of orientation, object represetation ambuigity issue 
Then, \textbf{we benchmark diverse object detection paradigms with AI-TOD-R} to investigate how different detection paradigms perform on oriented tiny objects (Section~\ref{sec:benchmark}). What distinguishes this benchmark from prior arts is that we break the fully-supervised paradigm and study both supervised and label-efficient methods, catering to broader and more practical applications. Our findings reveal that generic object detectors tend to exhibit abnormal results when confronted with extremely tiny objects.
Notably, the learning bias appears invariably across various methods. The objects' tiny size and low confidence characteristics make them easily suppressed or ignored during model training.
The vanilla optimization process will inevitably pose them into significantly \textit{biased prior setting} and \textit{biased sample learning} dilemmas, severely impeding the performance of oriented tiny object detection (Section~\ref{sec:unreveal}).
To address this issue, \textbf{we propose a new approach}: Dynamic Coarse-to-Fine Learning (DCFL), aimed at providing unbiased prior setting and sample supervision for oriented tiny objects (Section~\ref{sec:method}). On the one hand, we reformulate the static prior into an adaptively updating prior, thereby guiding more prior positions towards the main area of tiny objects. 
On the other hand, dynamic coarse-to-fine sample learning separates the label assignment into two steps: the coarse step offers diverse positive sample candidates for objects of various sizes and orientations, and the fine step warrants the high quality of positive samples for predictions. 
%The methodology part extends our conference version: DCFL. Similar to DCFL, the coarse step provides diverse and sufficient positive sample candidates for objects of various sizes and orientations, the fine step warrants the high quality of positive samples for predictions. Differently, we improve the Prior Capturing Block by introducing an explicit prior capturing target which guides more prior onto the object's main body, further mitigating oriented tiny objects' lack of semantically-rich positive samples. 

We perform experiments on eight heterogeneous benchmarks, including tiny/small oriented object detection (AI-TOD-R, SODA-A~\cite{soda_2023_pami}), oriented object detection with large numbers of tiny objects (DOTA-v1.5~\cite{DOTA2.0_2021_pami}, DOTA-v2~\cite{DOTA2.0_2021_pami}), multi-scale oriented object detection (DOTA-v1~\cite{DOTA_2018_CVPR}, DIOR-R~\cite{diorr_2022_tgrs}), and horizontal object detection (VisDrone~\cite{visdrone_2021_pami}, MS COCO~\cite{COCO_2014_ECCV}). Our results demonstrate that DCFL remarkably outperforms existing methods for detecting tiny objects (Section~\ref{sec:experiments}). Moreover, our results highlight three characteristics of DCFL: 
(1) \textbf{Costless improvement}: Extensive experiments on various datasets show that DCFL improves the detection performance without adding any parameter or computational overhead during inference. (2) \textbf{Versatility}: The DCFL approach can be plugged into both one-stage and two-stage detection pipelines and improve their performance on oriented tiny objects. Beyond oriented tiny objects, DCFL also enhances the detection performance of generic small objects. (3) \textbf{Unbiased learning}. By dissecting the training process, we reveal how DCFL achieves unbiased learning—adaptively updating priors to better align with tiny objects' main areas, while balancing the quantity and quality of samples across different scales.

%Extensive experiments on oriented tiny object detection, generic oriented object detection, and generic object detection scenarios demonstrate that our proposed method can significantly lift both one-stage and two-stage baselines by a large margin, quantitatively and qualitatively. Ablations verify the effectiveness of individual strategies and the robustness of parameter choices. In addition, we dissect the detailed behavior into the training process (prior, sample assignment) to help readers have a straightforward understanding of how the proposed architecture achieves unbiased learning.
% enlarge the feature resolution vs. dynamic prior
 
Aiming at addressing the challenging task of oriented tiny object detection, this paper provides a comprehensive extension of our previous conference version~\cite{dcfl_2023_cvpr}. Beyond methodological contributions published previously, this journal extension introduces several \textbf{additional advancements}:

\begin{itemize}
    \item Establishing a task-specific dataset for oriented tiny object detection, features the smallest object size among oriented object detection datasets, compensating for the lack of resources in this challenging area.
    \item Creating a benchmark that covers a variety of object detection paradigms, including both fully-supervised and label-efficient methods, revealing learning biases against oriented tiny objects across these approaches. 
    \item Demonstrating the versatility of DCFL by plugging it into both one-stage and two-stage methods, and verifying its generalization ability on small oriented object detection by validating on SODA-A dataset.
\end{itemize}

%this journal extension further introduces a dataset, benchmark, and enhances the experimental results via validating the proposed method on more datasets and providing more detailed analyses.

%method via improving the coarse-to-fine learning pipeline with an explicit prior capturing target, and perform extensive experiments on more datasets. 

%% file: relatedworks.tex
\section{Related works}

\subsection{Small and Oriented Object Detection Datasets}
\textbf{Small and tiny object detection datasets.} Due to the lack of specialized datasets, early studies on Small Object Detection (SOD) are mainly based on small objects in generic or some task-specific datasets.
For example, the generic object detection dataset MS COCO~\cite{COCO_2014_ECCV}, face detection dataset WiderFace~\cite{widerface_2016_cvpr}, pedestrian detection dataset EuroCity Persons~\cite{europerson_pami_2019}, and Drone-view dataset VisDrone~\cite{visdrone_2021_pami} all contain a considerable number of small objects that could assist related studies.
As the SOD performance has been struggling for a long time, the establishment of specialized dataset for SOD is receiving growing attention. TinyPerson~\cite{TinyPerson_2020_WACV} is the first dataset designed for tiny-scale person detection. AI-TOD~\cite{AI-TOD_2020_ICPR,aitodv2_2022_isprs} is the first multi-category dataset for tiny object detection. DTOD~\cite{densetiny_2024_tgrs} compounds the challenge by addressing not only the tiny size of objects but also their dense packing. Recently, the introduction of the first large-scale SOD dataset SODA~\cite{soda_2023_pami} along with its benchmark further highlights the necessity of targeted research on SOD.

\textbf{Oriented object detection datasets.} Oriented object detection is an important direction of visual detection since orientation information significantly reduces the background region in bounding boxes with minimal additional parameters. 
The multi-scale datasets DOTA-v1/1.5/2~\cite{DOTA_2018_CVPR,DOTA2.0_2021_pami} and DIOR-R~\cite{diorr_2022_tgrs} are widely adopted for performance benchmarking, where the DOTA-v2 is also characterized by its large number of small objects. In addition to these generic datasets, task-specific datasets are also introduced to dissect some targeted problems. For example, some datasets are established to study specific classes (\textit{e.g.}, HRSC2016~\cite{HRSC2016_2016}, UACS-AOD~\cite{UCAS_AOD_2015_ICIP}, VEDAI~\cite{VEDAI_2016_JVCIR}), some are designed for the fine-grained object detection (\textit{e.g.}, FAIR1M~\cite{fair1m_2022_isprs}), while some datasets are proposed for specific modalities (\textit{e.g.}, SSDD~\cite{SSDD_RS_2021}).
Meanwhile, there are also datasets designed for other scenarios sensitive to the object's orientation, including text~\cite{icdar_2015}, retail~\cite{sku110k_2019_cvpr}, and crack detection~\cite{iccv_2023_crack}. 

\subsection{Oriented Object Detection}

\textbf{Prior design.} The anchor mechanism is a classic prior design that can facilitate the training of both generic object detectors and oriented object detectors. 
As a pioneering work, rotated RPN~\cite{rotatedrpn_2018_tmm} extends horizontal anchors to the field of oriented object detection via presetting 54 anchors with various scales and angles for each feature point. 
Although this approach improves recall by covering a wide range of \textit{gt} shapes, it comes with an increased computational cost. 
Afterwards, RoI Transformer~\cite{RoI-Transformer_2019_CVPR} learns to transform RPN-generated horizontal proposals to oriented proposals, achieving more accurate feature alignment while simplifying the anchor design. 
Toward a simpler and more efficient framework, the Oriented R-CNN~\cite{orientedrcnn_2021_iccv} designs an oriented RPN that directly predicts oriented proposals based on horizontal anchors. 
More recently, one-stage oriented object detectors gradually emerged, including anchor-based detectors~\cite{R3Det_2021_AAAI,s2anet_2021_tgrs} with box-based prior and anchor-free detectors~\cite{fcosr_2021_arxiv,orientedrep_2022_cvpr,arsdetr_tgrs_2024} with point-based or query-based prior.

\textbf{Label assignment.} The label assignment process separates the prior positions into positive or negative learning samples, playing a pivotal role in object detection~\cite{paassignment_2020_eccv,ota_2021_cvpr,iqdet_2021_cvpr,atss_2020_cvpr}. 
In the field of oriented object detection, several methods have been proposed to enhance the quality of label assignment.
DAL~\cite{dal_2021_aaai} addresses the inconsistency between input prior IoU and output predicted IoU by defining a matching degree as a soft label that dynamically reweights the anchors. More recently, SASM~\cite{sasm_2022_aaai} introduces a shape-adaptive sample selection and measurement strategy, accurately assigning labels according to the object's shape and orientation.  
Similarly, GGHL~\cite{gghl_2022_tip} proposes fitting the main body of an instance with a single 2-D Gaussian heatmap, dividing and reweighting samples in a dynamic manner. In addition, Oriented Reppoints~\cite{orientedrep_2022_cvpr} improves the RepPoints~\cite{RepPoints_2019_ICCV} by assessing the quality of points, refining the detection performance.

\subsection{Tiny Object Detection}

\textbf{Sample learning.} 
Tiny objects usually suffer from low matching degrees with static anchors or limited coverage of feature point priors, resulting in a lack of positive samples. In generic object detection, the adaptive label assignment strategy ATSS~\cite{atss_2020_cvpr} implicitly reconciles the number of positive samples for objects of different scales. Explicitly targeting the sample learning issues of tiny objects, NWD-RKA~\cite{aitodv2_2022_isprs} and RFLA~\cite{rfla_2022_eccv} propose distribution-based similarity measurement and sample assignment strategies to achieve scale balanced learning. More recently, the CFINet~\cite{cfinet_iccv_2023} improve the detection performance of small objects by employing dynamic anchor selection and cascade regression to generate high-quality proposals.

\textbf{Feature enhancement.} Small or tiny objects themselves show very limited features, some studies thus propose to leverage external content to enhance the features of small objects with super-resolution or GAN. Among them, PGAN~\cite{PGAN_2017_CVPR} is the pioneering work that applies GAN to small object detection. Besides, Bai~\textit{et al.}~\cite{SOD-MTGAN_2018_ECCV}~introduce the MT-GAN which trains an image-level super-resolution model to improve the RoI features of small objects. CFINet~\cite{cfinet_iccv_2023} also enhances small objects' feature representation through mimicking high-quality features. Other notable methods that leverage super-resolution for small object detection include works such as ~\cite{Better_to_Follow_2019_ICCV,auxiliarygan_2021_rs,residualsuperres_2021_rs,edgegan_2020_rs}.

\textbf{Metric design.} Tiny objects often have a low tolerance for bounding box perturbation under generic location metrics like IoU. To address IoU-induced issues throughout the detection pipeline, specialized metrics have been designed to better handle tiny objects. To improve the label assignment performance, DotD~\cite{dotd_2021_cvprw}, series of works like NWD~\cite{nwd_2021_arxiv}, RFLA~\cite{rfla_2022_eccv}, and KLDet~\cite{kldet_2024_tgrs} introduce either center-based or distribution-based metrics. These approaches mitigate sample imbalance issues caused by overlap-based measurements. On the other hand, loss metrics designed to achieve scale-invariant~\cite{GIoU_loss_2019_CVPR,diou_2020_aaai}, evaluation consistent~\cite{Unitbox_2016_ACMM}, and boundary continuous~\cite{gaussian3d_2022_pami,psc_2023_cvpr} location regression also offer valuable insights into improving tiny object detection.

Despite these progress, the existing literature falls short in handling extremely oriented tiny objects. First, there still lacks a task-specific dataset and benchmark aiming at detecting the challenging but ubiquitous oriented tiny objects. Second, current detection paradigms cannot simultaneously manage the prior and sample biases in oriented tiny object detection, resulting in sub-optimal performance.
In this work, we aim to bridge these gaps by 1) further pushing the limits of object size in the dataset and benchmark for oriented object detection, and 2) proposing an unbiased prior update and sample learning pipeline that enables detectors to be supervised by more high-quality oriented tiny object samples during training.

%% file: aitodr.tex
\section{AI-TOD-R Dataset}
\label{sec:dataset}
\begin{figure*}[t]
\centering
\includegraphics[width=\linewidth]{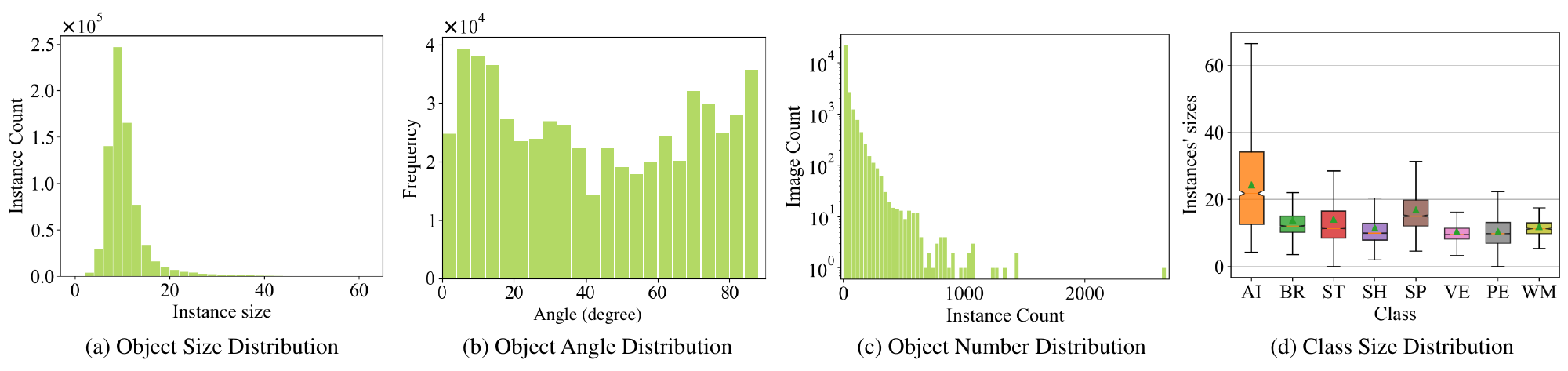}
\caption{Statistical analysis of the AI-TOD-R. From left to right, we show the dataset's object size distribution, object angle distribution, object number per image distribution, and class size distribution, respectively. The box plot of ``Class Size Distribution" shows the object's absolute size's mean value and standard deviation within each class. }
\label{aitodr_stat}
\end{figure*}

\input{dataset_size}

\subsection{Semi-automatic Annotation}
\begin{figure}[t]
\centering
\includegraphics[width=\linewidth]{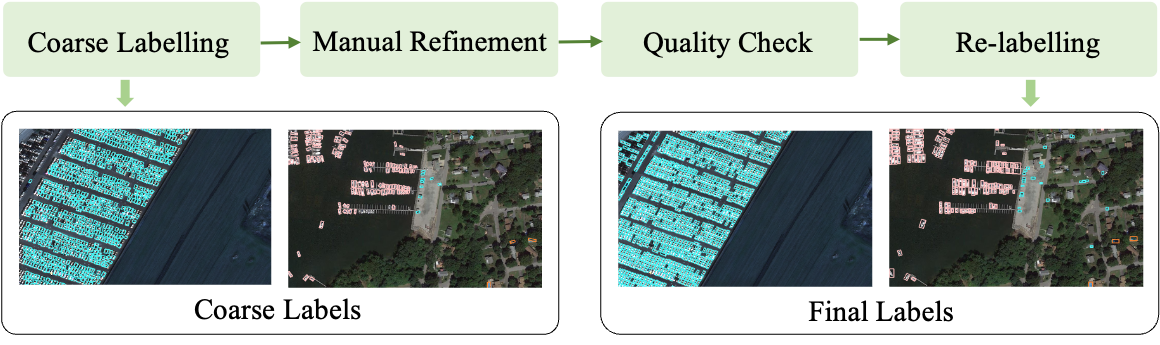}
\caption{The labelling process of the AI-TOD-R. The coarse labels are automatically generated by H2RBox-v2, and final labels are obtained by manual labelling and verification.}
\label{process}
\end{figure}

Due to weak features and large quantity, arbitrarily oriented tiny objects are easily confused with the background, making the artificial annotation process difficult and laborious. To guarantee high annotation quality and reduce annotation cost, we employ a semi-automatic annotation protocol that is composed of three basic steps: algorithm-based coarse labelling, manual refinement, and quality double-checking. The illustration of this process is shown in Figure~\ref{process}.

In the first step, we use the weakly supervised detector to generate OBB predictions with an HBB tiny object detection dataset.
Prior to this work, our AI-TOD-v2~\cite{aitodv2_2022_isprs}, characterized by its extremely tiny object size, multi-source images, and high-quality HBB annotation, lays the foundation for AI-TOD-R. 
Meanwhile, recently emerging weakly supervised methods are capable of predicting OBBs under only HBB supervision~\cite{h2rbox_2022_iclr,h2rboxv2_2024_nips}, achieving competitive performance with fully supervised methods on the DOTA dataset~\cite{DOTA_2018_CVPR,DOTA2.0_2021_pami}. Combining the merits of existing datasets and methods, we use the SOTA weakly supervised method H2RBox-v2~\cite{h2rboxv2_2024_nips} to generate OBB predictions based on the AI-TOD-v2 dataset, which serves as the initial annotations. 

In the second step, we manually refine the algorithm-generated preliminary annotations to fix errors. Although the current weakly supervised method can provide OBB predictions under HBB supervision, their performance on tiny objects remains far from satisfactory, particularly in scenarios with weak, densely packed objects. Existing algorithms tend to produce false negative and inaccurate predictions, as shown in Figure~\ref{process}. 
Consequently, we manually adjust the initial annotations to fix error annotations with the following approach. First, we select some typical images and call for experts to re-annotate them with the help of visual results from AI-TOD-v2, establishing an annotation guide. 
Based on this guide, we train volunteers with background in computer vision to perform large-scale adjustments. Volunteers are encouraged to adjust inaccurate predictions that they are confident with and mark down the image ID of cases that they are uncertain about. These uncertain cases are then resolved through team discussions and voting. 

Finally, we call for experts and volunteers to double-check each image and find out low-quality annotations. We then redistribute these images to volunteers for re-annotation. This combination of algorithmic initial annotation and meticulous manual refinement by a collaborative team ensures the high quality of the dataset.

\subsection{Statistical Analysis}

AI-TOD-R is currently the dataset of the smallest object size in the field of oriented object detection, containing a total of 8 classes, 28,036 images, and 752,460 objects with oriented bounding boxes and category labels. The dataset is divided into the {\tt train} set, {\tt val} set, {\tt trainval} set, and {\tt test} set. In the following, we present a comprehensive statistical analysis and a comparative evaluation of the characteristics of this data set against similar data sets.

\textbf{Extremely tiny object size.} As shown in Table~\ref{tab:dataset_scales}, the mean object size of AI-TOD-R is only 10.6$^{2}$ pixels, which is the smallest among all datasets. The detailed object size distribution is shown in Figure~\ref{aitodr_stat}(a), where most objects are gathered within the tiny scale ($\textless 16\times16$ pixels). Different from previous datasets proposed for generic oriented object detection or small oriented object detection, the extremely tiny mean object size and massive tiny objects make AI-TOD-R a challenging dataset dedicated to the oriented tiny object detection task.

\textbf{Arbitrary object orientations.} We employ the OpenCV definition to analyze the distribution of an object's rotation angles. The object's rotation angle is defined as the angle between the bounding box and the horizontal axis, with a range of $(0, 90^{\circ}]$. The dataset's object angle distribution is shown in Figure~\ref{aitodr_stat}(b). AI-TOD-R contains a large number of objects across various rotation angles, demonstrating its characteristic of arbitrary orientation. This feature aids detectors in learning object representation robust to different rotation angles.

\textbf{Massive objects per image.} In addition to the tiny scale and arbitrary orientations, another distinct characteristic of this dataset is the large quantity of objects in each image. Aerial imagery captures layout information with a broad field of view, resulting in the large number of objects covered by each image. According to Figure~\ref{aitodr_stat}(c), an image in AI-TOD-R can contain over 2000 objects, with most images featuring over 100 objects. The vast number of tiny objects in each image significantly increases the computational burden during training and inference, giving rise to the need for efficient detector designs that facilitate practical applications. 

\textbf{Imbalanced class distribution.} Like many generic object detection or oriented object detection datasets, the class imbalance challenge also exists in AI-TOD-R. This imbalance is reflected in the object number\footnote{airplane (1,667), bridge (1,541), storage-tank (13,771), ship (35,813), swimming-pool (1,617), vehicle (662,929), person (34,490), wind-mill (632)} and object size distribution (Figure~\ref{aitodr_stat}(d)) for each class. This imbalance depicts the real-world class distribution and also calls for robust oriented object detectors capable of class-balanced detection performance.

\subsection{Label Visualization}

\begin{figure*}[t]
\centering
\includegraphics[width=\linewidth]{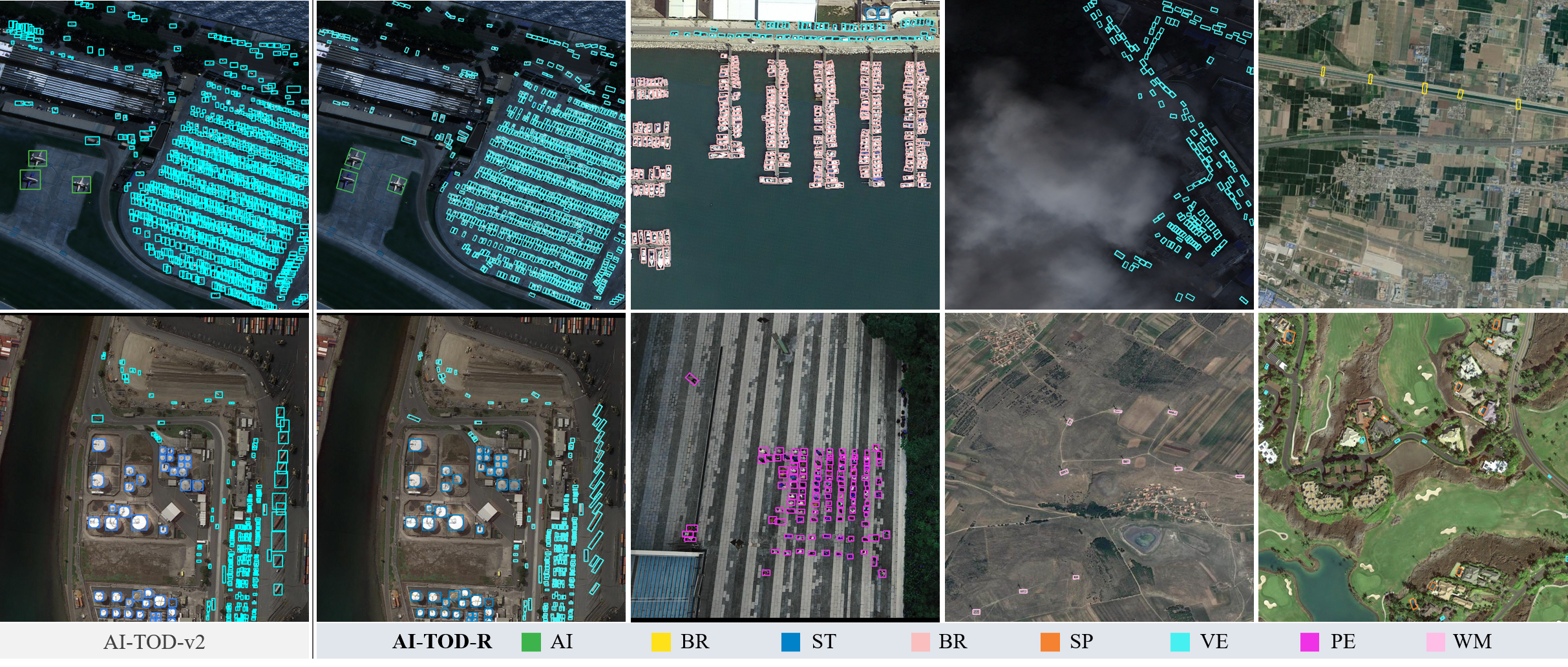}
\caption{Visualization of annotations in AI-TOD-R. Compared to AI-TOD-v2, using oriented bounding boxes to represent tiny objects can significantly reduce back noise, and this advantage is particularly obvious in densely arranged scenarios. In addition to the extremely tiny object size, AI-TOD-R introduces other challenges like dense arrangement, weak feature representation, and imbalanced class distributions.}
\label{aitodr_vis}
\end{figure*}

Figure~\ref{aitodr_vis} showcases typical samples from the AI-TOD-R dataset. These typical samples exhibit characteristics of the dataset, including extremely tiny object scale, arbitrary orientation, dense arrangement, and complex scenes. 
In particular, the visualized annotations reveal the unique advantages of representing tiny objects with oriented bounding boxes. Using oriented bounding boxes to represent objects allows the annotation boxes to more tightly enclose the object's main area. This advantage is particularly evident in densely packed regions, where rotated bounding boxes can significantly reduce overlap between adjacent object boundaries, thus preventing confusion during the network's learning and prediction processes in such areas. In addition, oriented bounding boxes can capture the orientation information of moving objects, such as vehicles in motion or ships at sea, providing richer information for downstream applications.

%% file: dataset_size.tex
\begin{table}
	\centering
	\caption{Comparison of image and object features in different object detection datasets. HBB, and OBB denote horizontal, and oriented bounding box, respectively. Object size is represented in the form of $mean\pm std$ of the dataset.}
    \resizebox{1.0\linewidth}{!}{
	\begin{tabular}{l|cccc}  
	\toprule
    Dataset  & Image Height & Image Count & Type & Object Size \\
	 \midrule
    MS COCO\cite{COCO_2014_ECCV} &   800-1333 & 163,957 & HBB & $99.5\pm 107.5$ \\
    DIOR\cite{DIOR_2019_ISPRS}   &   800 & 23,463 & HBB &$65.6\pm 91.7$   \\
    DIOR-R\cite{DIOR_2019_ISPRS}   &   800 & 23,463 & OBB &  $57.7\pm 80.2$ \\
    DOTA-v1.0\cite{DOTA_2018_CVPR} & 800-13000 & 2,423 & H/OBB & $55.3\pm 63.1$   \\
    VisDrone\cite{visdrone2019_2019_iccvw} & 2000 & 8,629 & HBB & $35.8\pm 32.8$   \\
    xView\cite{xview_2018_arXiv}   & -3000 & 1,127 & HBB & $34.9\pm 39.9$    \\
    DOTA-v1.5\cite{DOTA2.0_2021_pami}  & 800-13000 & 2,423 & H/OBB & $34.0\pm 47.8$    \\
    DOTA-v2\cite{DOTA2.0_2021_pami}  & 800-13000 & 11,268 & H/OBB &  $24.8\pm 32.6$    \\
    VEDAI (512)\cite{VEDAI_2016_JVCIR}  & 512, 1024 & 1,210 & HBB & $33.4\pm 11.3$    \\
    SODA-D\cite{soda_2023_pami}  & 3407 & 24,828 & OBB & $25.4\pm 10.0$  \\
    TinyPerson\cite{TinyPerson_2020_WACV}   & 1000-5616 & 1,610 & HBB & $18.0\pm 17.4$   \\
    SODA-A\cite{soda_2023_pami}  & 4761 & 2,513 & OBB & $15.6\pm 7.6$ \\
    AI-TOD-v2~\cite{aitodv2_2022_isprs}     & 800 & 28,036 & HBB & $12.7\pm 5.6$   \\
    \midrule
    AI-TOD-R  & 800 & 28,036 & OBB & $10.6\pm 4.9$\\
	\bottomrule
	\end{tabular}}
	\label{tab:dataset_scales}
\end{table}

%% file: benchmark.tex
\section{AI-TOD-R Benchmark}
\label{sec:benchmark}
In this section, we present a comprehensive benchmark for AI-TOD-R, encompassing fully-supervised oriented object detection methods as well as label-efficient methods consisting of semi-supervised object detection (SSOD), sparsely annotated object detection (SAOD), and weakly-supervised object detection (WSOD) methods.

\subsection{Implementation Details}
For fully-supervised methods, experiments on the AI-TOD-R are performed following the default setting of AI-TOD series~\cite{AI-TOD_2020_ICPR,aitodv2_2022_isprs}. We use AI-TOD-R's {\tt trainval set} for training and its {\tt test set} for evaluation, and retain the image size as $800\times800$ for training and testing. The batch size and learning rate are set to 2 and 0.0025 respectively. We only use random flipping as data augmentation for all experiments.

For label-efficient methods, we reorganize training labels and schedules to adapt to different paradigms. Semi-Supervised Object Detection (SSOD) methods randomly retain annotations with 10$\%$, 20$\%$, and 30$\%$ of the images from the AI-TOD-R's {\tt trainval set} as training annotations. We follow the default settings of SOOD~\cite{sood_cvpr_2023} with a batch size of 6 (with a 1:2 ratio of unlabeled to labeled data) and a learning rate of 0.0025. Additionally, we maintain the same total number of \textit{batch size $\times$ iterations} as the fully-supervised 40-epoch setup. Sparsely Annotated Object Detection (SAOD) randomly retains 10$\%$, 20$\%$, and 30$\%$ annotations of all objects from the AI-TOD-R's {\tt trainval set} as training labels. We use a batch size of 2, and a learning rate of 0.0025, and maintain the same total number of \textit{batch size $\times$ iterations} as the fully-supervised 40-epoch setup. Besides, Weakly Supervised Object Detection (WSOD) mainly switches the {\tt trainval set}'s annotations from OBB to HBB and keeps other settings as the fully-supervised setting. All other settings are retained as their baseline methods unless otherwise specified.

\input{aitodr_benchmark}

\subsection{Results of Fully-supervised Methods}

In Table~\ref{table:supervised}, we benchmark the detection performance on oriented tiny objects across a wide range of oriented object detectors. To better compare and analyze the characteristics of various detection paradigms on the oriented tiny object detection task, we introduce them in a classified manner.

\textbf{Basic architecture.} Based on the prior setting and stage number, oriented object detection architectures can be separated into \textit{dense}~\cite{Focal-Loss_2017_ICCV,FCOS_2022_TPAMI} (\#1, 2),  \textit{dense-to-sparse}~\cite{Faster-R-CNN_2015_NIPS,RoI-Transformer_2019_CVPR,orientedrcnn_2021_iccv} (\#3, 4, 5), and \textit{sparse}~\cite{deformabledetr_2021_iclr,arsdetr_tgrs_2024} (\#6, 7) paradigms. The \textit{dense} paradigm usually refers to one-stage methods that yield dense predictions per feature point, \textit{dense-to-sparse} methods use the first stage to generate sparse proposals (\textit{e.g.}, RPN) and refine proposals as final predictions in the second stage (\textit{e.g.}, R-CNN), while the \textit{sparse} paradigm is mainly based on Transformers to reason about the object's class and location with a set of sparse queries. 
Among the \textit{dense} paradigm, FCOS-O releases the IoU-constrained assignment by labeling \textit{gt}-covered points as positive samples, performing better than the anchor-based \textit{dense} method. Benefiting from the FPN with higher resolution (\textit{P2}) and feature interpolated RoI Align, \textit{dense-to-sparse} methods perform slightly better than \textit{dense} methods, while at the cost of higher computation demand. Compared to other paradigms, the state-of-the-art \textit{sparse} method (\#7) gradually performs favorably on oriented tiny objects, mainly attributed to its training strategies tailored from advanced generic detectors and its rotated deformable attention optimized for arbitrary-oriented objects.

\textbf{Box representation and loss design.} The vanilla regression-based loss suffers from issues including inconsistency with evaluation metrics, boundary discontinuity, and square-like problems, giving rise to numerous box representation studies. Here, oriented tiny object detection also benefits from these improved representations and their induced loss functions. The Gaussian-based loss~\cite{kld_2021_nips,kfiou_2022_iclr} (\#8, 9) eradicates the boundary discontinuity issue and enforces the alignment between the optimization goal with the evaluation metric, slightly improving the AP for about 1 point based on the RetinaNet-O baseline. Notably, the point set-based method~\cite{orientedrep_2022_cvpr,sasm_2022_aaai,beyond_2021_cvpr} (\#10, 16, 17) is particularly effective for detecting oriented tiny objects, which may be attributed to the deformable points' representation robustness to extreme geometric characteristics.

\textbf{Sample selection strategies.} The quality of positive sample selection directly affects the supervision information in the training process, playing a crucial role in tiny object detection. By adaptively determining the positive anchor threshold for each \textit{gt}, ATSS-O lifts the RetinaNet-O baseline by 3.6 points. By dynamically assessing the sample quality based on the object's arrangement and shape information, CFA and SASM yield promising performances of 11.4\% and 12.4\%, respectively. These significant improvement raised by sample selection strategies (\#15-17) further highlights the importance of customized sample assignment methods for oriented tiny objects.

\textbf{Backbone choice.} We analyze the effects of various backbones on oriented tiny object detection by investigating \textit{deeper architecture}, \textit{vision transformer}, \textit{large convolution kernels}, and \textit{rotation equivariance}. Different from generic object detection, oriented tiny object detection does not benefit from deeper backbone architecture (\#18 \textit{vs.} \#5) or large convolution kernels (\#20 \textit{vs.} \#5), where these improved backbones retain similar AP with the basic ResNet-50. This interesting phenomenon can be largely attributed to the limited and local information representation of tiny objects. After multiple times of down-sampling in deeper layers of the network, the limited information of tiny objects is further lost. Besides, the large receptive field of large convolution kernels struggles to fit or converge to the extremely tiny region of interest.
By contrast, the shifted window transformer (Swin Transformer~\cite{swin_2021_iccv}) and rotation-equivalent feature extraction (ReDet~\cite{redet_2021_cvpr}) could also benefit oriented tiny objects based on our experiments. 

\subsection{Results of Label-efficient Methods}

Label-efficient object detection aims at simplifying the annotation cost (\textit{e.g.}, quantity, difficulty), meanwhile aligning or even surpassing the performance with fully-supervised methods. Label-efficient approaches show great demand and potential on oriented tiny objects since their annotation process is quite laborious and difficult. Herein, we investigate three kinds of dominant label-efficient paradigms as follows, whose results on the AI-TOD-R dataset are listed in Table~\ref{tab:labelefficient}.

\input{aitodr_labelefficient}

\textbf{Semi-Supervised Object Detection (SSOD).} SSOD relieves the annotation burden via leveraging the precious annotated images and massive unlabelled images to train object detectors efficiently. Current state-of-the-art approaches~\cite{unbiased_2021_iclr,softteacher_2021_cvpr,sood_cvpr_2023} employ a teacher-student network architecture with a pseudo-labelling fashion. Surprisingly, using only 30\% labelled images and the remaining unlabelled images, the state-of-the-art SSOD approach: SOOD~\cite{sood_cvpr_2023} (40 epochs) has already achieved competitive performance with fully-supervised single-stage counterparts (\textit{i.e.}, FCOS-O with 1$\times$) using full-set annotation. 
The uncovers the great potential and application value of SSOD methods on tiny-scale oriented objects.

\textbf{Sparsely Annotated Object Detection (SAOD).} SAOD approaches propose to randomly annotate a proportion of objects throughout the whole training set for label-efficient learning. We adapt a classic SAOD method to oriented tiny object detection (\textit{i.e.}, Co-mining~\cite{comining_2021_aaai}). Despite using the same number of annotated objects, SSOD methods outperform the SAOD method tested. This performance gap may be attributed to the fact that Co-mining does not utilize the advanced teacher-student network, thereby limiting its effectiveness.

\textbf{Weakly-Supervised Object Detection (WSOD).} Another popular direction of label-efficient object detection uses coarse-level annotations, which are more easily accessible, for fine-level predictions. Among them, a dominant line of research lies in using horizontal bounding box supervision for oriented box prediction (\textit{e.g.}, H2RBox~\cite{h2rbox_2022_iclr}). With advanced training strategies, experiments reveal that merely using HBB supervision has shown comparable performance with OBB-supervised single-stage baselines (\textit{e.g.}, FCOS-O).

In short, label-efficient methods have demonstrated excellent performance in the task of oriented tiny object detection. Training with much fewer annotations, SSOD and WSOD methods show very competitive performance compared to one-stage fully-supervised baselines. These findings demonstrate the significant application value and potential for further exploration of label-efficient methods in the field of oriented tiny object detection.

\begin{figure}[t]
\centering
\includegraphics[width=\linewidth]{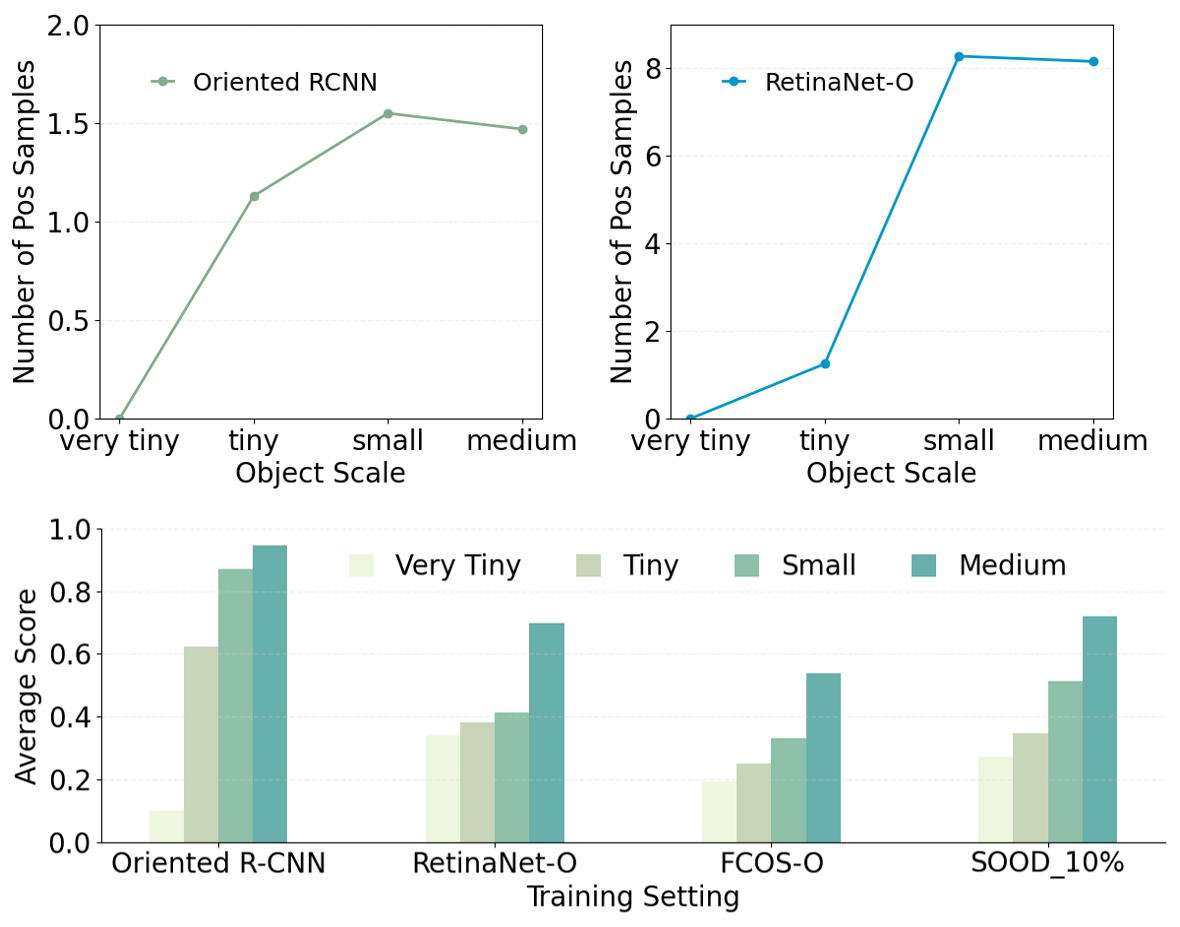}
\caption{An illustration of the sample learning bias. SOOD~\cite{sood_cvpr_2023} is trained with 10\% labels under the semi-supervised object detection pipeline.}
\label{posterior_bias}
\end{figure}

\subsection{Uncovering Learning Bias}
\label{sec:unreveal}

Despite the differences in detection paradigms, one consistent finding is that the detection performance of oriented tiny objects remains significantly inferior to that of regular-sized objects. To gain a clearer understanding of the underlying reasons for this performance gap, we conduct a statistical analysis from the perspective that directly drives the model's training: the sample learning process of objects across various scales (\textit{i.e.}, the input sample and output predictions).

Specifically, we investigate the prior matching degree (input) and posterior confidence scores (output) for different-sized objects when training. 
The results, presented in Figure~\ref{posterior_bias}, show the prior sample selection results across different detectors by counting the number of positive samples assigned to objects of varying scales (upper line charts), and the model's posterior confidence scores for different-sized objects (lower bar chart).

Our analysis in Figure~\ref{posterior_bias} shows that \textit{oriented tiny objects often face a biased dilemma across various detection pipelines}. At the prior level, tiny-scale objects receive significantly fewer positive samples than larger-scale objects. This phenomenon can largely be attributed to the limited feature map resolution, sub-optimal measurement, and label assignment strategies.
Specifically, the stride between adjacent feature points and their corresponding prior locations (\textit{e.g.}, anchor box/point) is constrained by the feature map resolution. For example, the stride of prior locations in a typical single-stage detector is at least 8 pixels. This sparse and fixed prior setting fundamentally limits the number of sample candidates for tiny objects compared to larger ones, leading to a \textit{biased prior setting}. 
Furthermore, oriented tiny objects often have a lower similarity with the sparse prior boxes (\textit{e.g.}, RetinaNet-O~\cite{Focal-Loss_2017_ICCV}) or cover very few prior points (\textit{e.g.}, FCOS-O~\cite{FCOS_2022_TPAMI}), which exacerbates the problem. 
Under the generic sample selection strategies (\textit{e.g.}, MaxIoU, Center Sampling), the number of positive samples ultimately assigned to tiny objects is further reduced, leading to a serious \textit{sample bias problem}.

This learning bias against oriented tiny objects is also reflected in their high uncertainty levels in posterior predictions, as shown in the lower part of Figure~\ref{posterior_bias}. The low confidence scores can further exacerbate the learning bias against oriented tiny objects. In supervised learning, some methods propose to select or re-weight confident samples~\cite{ota_2021_cvpr,autoassign_2020_arxiv,paassignment_2020_eccv}, which will further weaken oriented tiny objects due to their high uncertainty levels.
In label-efficient learning, thresholds based on predicted scores are used to select pseudo-labels.
This size-induced bias will also be amplified in this process, as regular objects, having higher posterior confidence scores, are more likely to be selected as pseudo-labels for training, whereas tiny-sized objects are ignored and lack training, thereby widening the gap between tiny objects and regular objects.

This investigation naturally raises the question: \textit{can we develop a method that achieves unbiased learning for different objects?} The following section addresses this question by introducing a new learning pipeline composed of a dynamically updated prior setting and a dynamic coarse-to-fine sample selection scheme. Our prior setting breaks the limit of fixed prior position by adapting prior initialization to the object's main area, and our sample selection method improves the assignment rule by providing more and higher-quality positive samples for training oriented tiny objects.

%% file: aitodr_benchmark.tex
\setlength{\tabcolsep}{3pt}
\begin{table*}[t]
\centering
\caption{Main results of fully-supervised methods on AI-TOD-R. For the training schedule, 1$\times$ denotes 12 epochs and 40e denotes 40 epochs. Methods with ``-O'' mean the rotated version of base detectors, and the name in ``()'' denotes the baseline method.}
\label{table:supervised}
\renewcommand{\arraystretch}{0.85}
\renewcommand{\tabcolsep}{1.0mm}
\resizebox{0.99\linewidth}{!}{
\begin{tabular}{l|l|c|c|ccc|cccc|c}
	\toprule
	ID &Method & Backbone & Schedule & AP & $\rm{AP_{0.5}}$ & $\rm{AP_{0.75}}$ & $\rm{AP_{vt}}$ & $\rm{AP_{t}}$ & $\rm{AP_{s}}$ & $\rm{AP_{m}}$ & \#Params.  \\
	\midrule
    \#\# & \textit{Architecture:} & & & & & & & & & & \\
    \#1  &  RetinaNet-O~\cite{Focal-Loss_2017_ICCV}   & ResNet-50 & 1$\times$ 	& 7.3  & 23.9 & 1.8 & 2.2 & 5.9 & 11.1 & 15.4 & 36.3M  \\
    \#2  &  FCOS-O~\cite{FCOS_2022_TPAMI} & ResNet-50 & 1$\times$  &  11.0 & 33.6	& 3.7 & 3.0 & 8.9 & 15.7 & 22.0 & 31.9M \\
    \#3  &  Faster R-CNN-O~\cite{Faster-R-CNN_2015_NIPS}    & ResNet-50 & 1$\times$ & 10.2	& 30.8 & 3.6 & 0.6 & 7.8 & 19.0 & 22.9  & 41.1M \\ 
    \#4 &  RoI Transformer~\cite{RoI-Transformer_2019_CVPR}    & ResNet-50 & 1$\times$  & 10.5	& 34.0 & 2.2 & 1.1 & 8.8 & 16.9 & 20.3 & 55.1M\\
    \#5 &  Oriented R-CNN~\cite{orientedrcnn_2021_iccv}    & ResNet-50 & 1$\times$  & 11.2 & 33.2	& 4.3 & 0.6 & 9.1 & 19.5 & 23.2 & 41.1M  \\
    \#6  &  Deformable DETR-O~\cite{deformabledetr_2021_iclr}   & ResNet-50 & 1$\times$  & 8.4 & 26.7 & 2.0 & 4.8 & 9.3  & 8.6 &  7.3 & 40.8M\\
    \#7 &  ARS-DETR~\cite{arsdetr_tgrs_2024}  & ResNet-50 & 1$\times$  & 14.3 & 41.1 & 5.8 & 6.3 & 14.5 & 17.6 & 18.7 &	41.1M \\
    \midrule
    \#\# & \textit{Representation:} & & & & & & & & & & \\
    \#8  &  KLD (RetinaNet-O)~\cite{kld_2021_nips} &  ResNet-50 & 1$\times$ & 7.8  & 24.8 & 2.3 & 3.1 & 6.7 & 10.3 & 15.8 & 36.3M   \\   
    \#9  &  KFIoU (RetinaNet-O)~\cite{kfiou_2022_iclr} & ResNet-50 & 1$\times$  &  8.1 & 25.2 & 2.8 & 2.0 & 6.6 & 12.3 & 17.1 & 36.3M \\ 
    \#10  &  Oriented RepPoints~\cite{orientedrep_2022_cvpr}  & ResNet-50 	& 1$\times$  & 13.0	& 40.3 & 4.2 & 5.2 & 12.2 & 16.8 & 21.4 & 36.6M \\
    \#11  &  PSC (RetinaNet-O)~\cite{psc_2023_cvpr}  & ResNet-50 & 1$\times$   & 	4.5 & 15.8 & 1.2 & 1.0 & 3.7 & 8.2 & 12.7 & 36.4M\\ 
    \#12  &  Gliding Vertex~\cite{Gliding_Vertex_2020_PAMI}  & ResNet-50 & 1$\times$  & 8.1 & 27.4	& 2.1 & 0.9 & 6.7 & 14.7 & 17.9 &	41.1M \\
    \midrule
    \#\# & \textit{Refinement:} & & & & & & & & & & \\
    \#13  &  $\rm{R^3Det}$~\cite{R3Det_2021_AAAI}  & ResNet-50 & 1$\times$  & 8.1	& 25.8 & 2.2 & 1.9 & 7.3 & 12.0 & 16.8 &  41.7M \\
    \#14  &  $\rm{S^2A}$-Net~\cite{s2anet_2021_tgrs}  & ResNet-50 	& 1$\times$   & 10.8 & 33.4 & 3.3  & 4.3 & 11.2 & 13.0 & 16.0 & 38.6M\\
    \midrule
    \#\# & \textit{Assignment:} & & & & & & & & & & \\
    \#15  &  ATSS-O (RetinaNet-O)~\cite{atss_2020_cvpr} & ResNet-50 	& 1$\times$  & 10.9	& 33.8 & 3.1 & 2.7 & 8.9 & 15.5 & 19.4 &  36.0M \\ 
    \#16 &  SASM \textcolor{darkgray}{(RepPoints-O)}~\cite{sasm_2022_aaai}  & ResNet-50 & 1$\times$  & 	11.4 & 35.0 & 3.7 & 3.6 & 10.2 & 15.4 & 19.8 &	36.6M \\
    \#17  &  CFA~\cite{beyond_2021_cvpr}  & ResNet-50 	& 1$\times$  & 12.4  & 38.7 & 4.0 & 5.0 & 11.9 & 16.5 & 18.8 & 36.6M \\
    \midrule
    \#\# & \textit{Backbone:} & & & & & & & & & & \\
    \#18  &  Oriented R-CNN   & ResNet-101 & 1$\times$  &	11.2 & 33.0	& 4.1 & 0.5 & 8.9 & 19.8 & 24.4 & 60.1M   \\
    \#19  &  Oriented R-CNN  & Swin-T & 1$\times$ &	12.0 & 34.6	& 4.6 & 0.7 & 9.9 & 20.8 & 25.3 & 44.8M   \\
    \#20  &  Oriented R-CNN   & LSKNet-T & 1$\times$  & 11.1 & 33.4 & 3.8 & 0.6 & 9.2 & 18.9 & 22.6 & 21.0M   
      \\
    \#21  &  ReDet~\cite{redet_2021_cvpr}  & ReResNet-50 & 1$\times$  & 11.6 & 32.8	& 4.8 & 1.4 & 9.5 & 19.4 & 23.2 & 31.6M  \\
	\midrule
    \midrule
    \#22  & DCFL (RetinaNet-O) & ResNet-50 	& 1$\times$  & 12.3 (+5.0) & 36.7 (+12.8) & 4.5 (+2.7) & 4.3 & 10.7 & 17.2 & 22.2 & 36.1M\\
    \#23  & DCFL (RetinaNet-O) & ResNet-50 	& 40e  & 15.2 (+7.9) & 44.9 (+21.0) & 5.1 (+3.3) & 4.9 & 13.1 & 19.7 & 25.9 & 36.1M\\
	\#24 & DCFL (Oriented R-CNN) & ResNet-50 	& 1$\times$  & 15.7 (+4.5) & 47.0 (+13.8) & 5.8 (+1.5) & 6.3 & 14.8 & 19.6 & 22.5 &  41.1M \\
	\#25 & DCFL (Oriented R-CNN) &  ResNet-50 	& 40e  & 17.1 (+5.9) & 49.0 (+15.8) & 7.2 (+2.9) & 6.4 & \textbf{16.0} & 21.6 & 24.9 &  41.1M \\
     \#26 & DCFL ($\rm{S^2A}$-Net) &ResNet-50 	& 1$\times$ & 13.7 (+2.9) & 39.7 (+6.3) & 5.3 (+2.0) & 4.7 & 12.4 & 18.6 & 22.6  & 38.6M \\
    \#27 & DCFL ($\rm{S^2A}$-Net) &ResNet-50 	& 40e & \textbf{17.5} (+6.7) & \textbf{49.6} (+16.2) & \textbf{7.9} (+4.6) & \textbf{6.5} & 15.7 & \textbf{22.6} & \textbf{27.4}  & 38.6M \\    
	\bottomrule
	\end{tabular}
 }
\end{table*}
\setlength{\tabcolsep}{1.4pt}
% ro to: rotated reppoints, lsknet-s, CSL
% SCHEDULE 1X, 3X

%% file: aitodr_labelefficient.tex
\begin{table*}[t!]
	\caption{Main results of label-efficient methods on AI-TOD-R. Evaluations are performed on the {\tt test set} of AI-TOD-R by training under different ratios of Oriented Bounding Box (OBB) annotations or Horizontal Bounding Box (HBB) annotations from its {\tt trainval set}. SSOD, SAOD, and WSOD denote semi-supervised object detection, sparsely annotated object detection, and weakly supervised object detection, respectively.}
	\label{tab:labelefficient}
    \centering
	\resizebox{0.99\textwidth}{!}{%
		\begin{tabular}{l|c|c|cccc|cccc|cccc|cccc}
			\toprule
			\multirow{2}{*}{Method} & \multirow{2}{*}{Category} & \multirow{2}{*}{Backbone} & \multicolumn{4}{c|}{10$\%$ OBB} & \multicolumn{4}{c|}{20$\%$ OBB} & \multicolumn{4}{c|}{30$\%$ OBB} & \multicolumn{4}{c}{100$\%$ HBB}                \\ \cmidrule(l){4-19} 
			&                           &  & AP & $\rm{AP_{0.5}}$ & $\rm{AP_{vt}}$ & $\rm{AP_{t}}$ & AP & $\rm{AP_{0.5}}$ & $\rm{AP_{vt}}$ & $\rm{AP_{t}}$ & AP & $\rm{AP_{0.5}}$ & $\rm{AP_{vt}}$ & $\rm{AP_{t}}$ & AP & $\rm{AP_{0.5}}$ & $\rm{AP_{vt}}$ & $\rm{AP_{t}}$ \\ \midrule
			Unbiased Teacher~\cite{unbiased_2021_iclr} & SSOD & ResNet-50 & 7.6 & 24.7 & 0.4 & 6.0 & 8.1 & 24.7 & 0.5 & 6.1 & 8.1 & 25.4 & 0.4 & 6.1 & - & - & - & -\\
			Soft Teacher~\cite{softteacher_2021_cvpr}         & SSOD & ResNet-50 & 9.4 & 29.0 & 0.3 & 7.6 & 10.2 & 31.1 & 0.5 & 7.9 & 10.4 & 32.2 & 0.6 & 7.8 & - & - & - & - \\ 
			SOOD~\cite{sood_cvpr_2023}         & SSOD & ResNet-50 & 9.4 & 29.3 & 2.8 & 8.1 & 12.1 & 35.7 & 3.5 & 10.2 & 13.0 & 38.8 & 3.9 & 11.1 & - & - &  - & - \\ 
			Co-mining~\cite{comining_2021_aaai}         & SAOD & ResNet-50 & 6.4 & 20.4 & 0.5 & 4.4 & 8.0 & 24.1 & 0.4 & 6.1 & 8.2 & 25.0 & 0.4 & 6.8 & - & - & - & - \\ 
			H2R-Box~\cite{h2rbox_2022_iclr}         & WSOD & ResNet-50 & - & - & - & - & - & - & - & - & - & - & - & - & 11.4 & 39.1 & 3.4 & 9.4 \\ 
			H2R-Box-v2~\cite{h2rboxv2_2024_nips}        & WSOD & ResNet-50& - & - & - & - & - & - & - & - & - & - & - & - & 11.7 & 38.2 & 4.6 & 9.5 \\
            \bottomrule
		\end{tabular}
	}
\end{table*}

%% file: method.tex
\section{Method}
\label{sec:method}
\begin{figure*}[t]
\centering
\includegraphics[width=\linewidth]{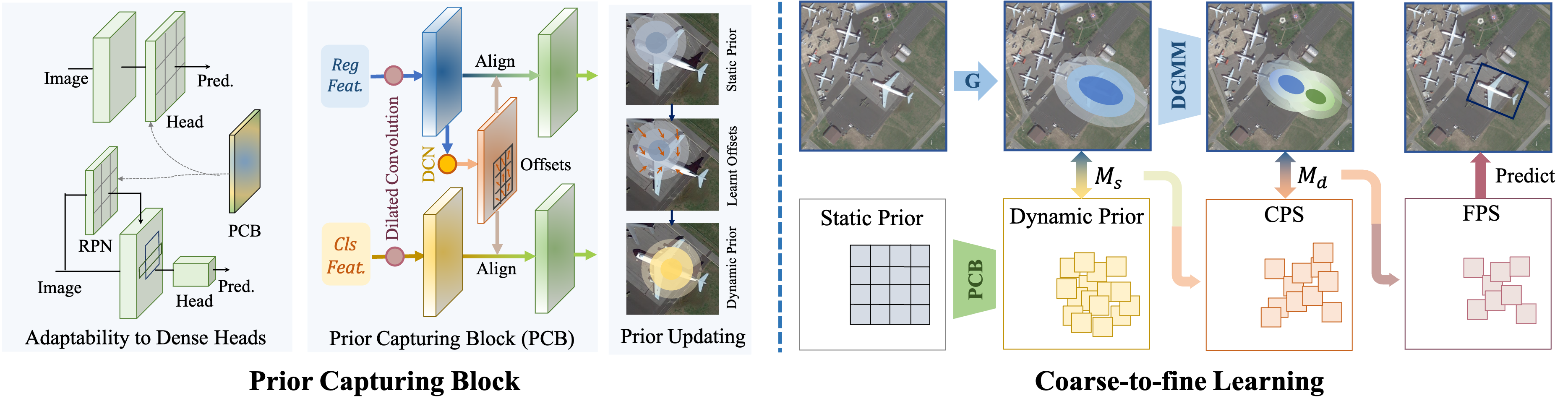}
\caption{An overview of the proposed method. The proposed DCFL learning scheme can be adapted into both one-stage and two-stage detection pipelines for oriented tiny object detection. Left: Feature extraction process and the prior updating process of the PCB. Right: The schematic diagram of the dynamic coarse-to-fine sample learning.}
\label{pipeline}
\end{figure*}

In this section, we first provide a paradigmatic comparison of our method with prior arts. Following this, we describe the details for core components (\textit{i.e.}, Dynamic Prior, Coarse Prior Matching, and Finer Posterior Matching) in our proposed DCFL. Figure~\ref{pipeline} shows an overview of the proposed method.

\subsection{Pipeline Overview}

\textbf{Static prior $\rightarrow$ Dynamic prior.} Oriented object detection is predominantly solved with \textit{dense} one-stage detectors (\textit{e.g.}, RetinaNet-O) or \textit{dense-to-sparse} two-stage detectors (\textit{e.g.}, Oriented R-CNN) nowadays~\cite{sparsercnn_2021_cvpr}. 
Different as architectures, their detection processes all initialize from a set of dense priors $P \in \mathbb{R}^{W \times H \times C}$ ($W \times H$: the size of the feature map, $C$: the number of prior information per feature point) and remap the set into final detection results $D$ through a Deep Neural Network (DNN), which can be simplified as:
\begin{equation}
    D = \mathrm{DNN}_{d}(P),
    \label{remap}
\end{equation}
where $\mathrm{DNN}_{d}$ is composed of the backbone and detection head. Detection results $D$ can be mainly separated into two parts: classification scores $D_{cls} \in \mathbb{R}^{W \times H \times A}$ ($A$ denotes the class number) and box locations $D_{reg} \in \mathbb{R}^{W \times H \times B}$ ($B$ is the box parameter number).

This static prior modeling suffers from significant prior bias issues for tiny objects: the prior position mostly deviates from the objects' main body (Section~\ref{sec:intro}). To accommodate the extreme sizes and arbitrary geometries of these tiny objects, we incorporate an iterative updating process for the prior position and refine it dynamically with each iteration. This transforms the prior into a dynamic set $\Tilde{P}$ (~$\Tilde{}$ denotes the dynamic item), leading to a reformulated detection process:
\begin{equation}
    D = \mathrm{DNN}_{d}(\underbrace{\mathrm{DNN}_{p}(P)}_{\text{Dynamic Prior}~\Tilde{P}}), 
\end{equation}
$\mathrm{DNN}_{p}$ is a learnable block incorporated within the detection pipeline to update the prior. 

\textbf{Static sample learning $\rightarrow$ Dynamic coarse-to-fine sample learning.} 
To train the $\mathrm{DNN}_{d}$, a proper matching between the prior set $P$ and the \textit{gt} set $GT$ needs to be solved to assign \textit{pos/neg} labels to $P$ and supervise the network learning. Existing assignment strategies can be classified into static and dynamic strategies.
For static assignment (\textit{e.g.}~RetinaNet~\cite{Focal-Loss_2017_ICCV}), the set of \textit{pos} labels $G$ is obtained via a hand-crafted matching function $\mathcal{M}_s$, and the set for a specific image remains the same for each epoch, which is formulated as: 
\begin{equation}
    G = \mathcal{M}_{s}(P, GT),
    \label{static_mapping}
\end{equation}
while dynamic assignment approaches~\cite{paassignment_2020_eccv,ota_2021_cvpr,dal_2021_aaai} tend to leverage the prior information $P$ along with posterior information (predictions) $D$ for dynamic sample selection, where they apply a prediction-aware mapping $\mathcal{M}_d$ to get the set $G$:
\begin{equation}
     G = \mathcal{M}_{d}(P, D, GT),
    \label{dynamic_mapping}   
\end{equation}
after the \textit{pos/neg} label separation, the loss function can be summarized into two parts:
\begin{equation}
    \mathcal{L}= \sum_{i=1}^{N_{pos}} \mathcal{L}_{pos}(D_{i}, G_{i}) +  \sum_{j=1}^{N_{neg}} \mathcal{L}_{neg}(D_{j}, y_j),
    \label{basic_loss}
\end{equation}
where $N_{pos}$, $N_{neg}$ are the number of positive and negative samples respectively, $y_j$ denotes the negative label.

Whether dynamic or static, oriented tiny objects are amidst a sample bias dilemma under existing label assignment methods: these strategies typically sample and weight high-scoring samples (\textit{i.e.}, prior location) as positive samples, while both prior and posterior scores for tiny objects are extremely low, making their effective samples wrongly labeled as outlier negative samples.

Towards unbiased sample learning, we reformulate this process into a dynamic coarse-to-fine learning pipeline based on the dynamic priors. The coarse step works in an object-centric way, where we construct a coarse positive candidate bag to warrant sufficient and diverse positive samples for each object.
The fine step aims at guaranteeing the learning quality, where we fit each \textit{gt} with a Dynamic Gaussian Mixture Model (DGMM) as a constraint to select high-quality samples.
Thus, the assignment process can be expressed as follows:
\begin{equation}
    \Tilde{G} = \mathcal{M}_{d}(\mathcal{M}_s(\Tilde{P}, GT), \Tilde{GT}),
    \label{c2f_mapping}
\end{equation}
the $\Tilde{GT}$ is a finer representation of an object with the DGMM. In a nutshell, our final loss is modeled as:
\begin{equation}
    \mathcal{L}= \sum_{i=1}^{\Tilde{N}_{pos}} \mathcal{L}_{pos}(\Tilde{D}_{i}, \Tilde{G}_{i}) +  \sum_{j=1}^{\Tilde{N}_{neg}} \mathcal{L}_{neg}(\Tilde{D}_{j}, y_j).
    \label{new_loss}
\end{equation}

\subsection{Dynamic Prior}
We introduce a dynamic updating mechanism that can benefit both \textit{dense} and \textit{dense-to-sparse} oriented object detection paradigm, named Prior Capturing Block (PCB). Seamlessly embedded into the original detection head, the PCB generates prior positions that are better aligned with the main body and geometries of tiny objects, increasing the number of high-quality sample candidates for these objects and mitigating the biased prior configuration.

The structure of the proposed PCB is illustrated in Figure~\ref{pipeline}. In this design, a dilated convolution is deployed to incorporate the object's surrounding context information, followed by the offsets prediction~\cite{DCN_2017_CVPR} to capture dynamic prior positions. 
Besides, the learned offsets from the regression branch are used to guide feature extraction in the classification branch, leading to better alignment between the two tasks. As such, the PCB inherits the flexibility of learnable priors in query-based detectors (\textit{e.g.}, DETR~\cite{detr_2020_eccv}) and retains the explicit physical meaning of static priors in dense detectors (\textit{e.g.}, RetinaNet-O).

The dynamic prior capturing process further unfolds as follows. As initialization, each prior location $\mathbf{p}(x, y)$ is set to the spatial location $\mathbf{s}$ of each feature point, which has been remapped to the image. In each iteration, we forward the network to capture the offset sets $\Delta\mathbf{o}$ of each prior location. Hence, the prior's location can be updated by:
\begin{equation}
    \Tilde{\mathbf{s}} = \mathbf{s} +st \sum_{i=1}^{n} \Delta \mathbf{o}_i/ 2n,
    \label{update_prior}
\end{equation}
where $st$ represents the stride of the feature map and $n$ is the vector number of offsets for each location. 

As a model-agnostic approach, the dynamic prior can be adapted into both one-stage and two-stage methods. 
More specifically, we use a 2-D Gaussian distribution $\mathcal{N}_p(\boldsymbol{\mu}_p, \boldsymbol{\Sigma}_p)$, which has proven conducive to small objects~\cite{rfla_2022_eccv,gwd_2021_icml} and oriented objects~\cite{gwd_2021_icml,kld_2021_nips},  to fit the prior's spatial location. Each dynamic prior location $\Tilde{\mathbf{s}}$ serves as the Gaussian mean vector $\boldsymbol{\mu}_p$, and each prior is associated with a square-shaped prior $(w, h, \theta)$ as their baseline detector, this shape information serves as the covariance matrix $\boldsymbol{\Sigma}_p$~\cite{gaussian3d_2022_pami}:

\begin{equation}\small
    \mathbf{\Sigma}_p=\begin{bmatrix}
        \cos{\theta} & -\sin{\theta} \\
        \sin{\theta} & \cos{\theta}
    \end{bmatrix}\begin{bmatrix}
    \frac{w^2}{4} & 0 \\ 
    0 & \frac{h^2}{4}
    \end{bmatrix}\begin{bmatrix}
        \cos{\theta} & \sin{\theta} \\
        -\sin{\theta} & \cos{\theta}
    \end{bmatrix}.
    \label{gaussian_model}
\end{equation}

\subsection{Dynamic Coarse-to-Fine Learning}
Without specialized consideration of tiny-scale objects, previous sample assignment strategies are biased towards sampling large object samples which usually hold higher confidence, discarding tiny-scale oriented objects as background. Towards scale-unbiased optimization, we design a dynamic coarse-to-fine learning pipeline, where the coarse step offers sample diversity while the fine step warrants learning quality.

\textbf{Coarse prior matching for sample diversity}. In the coarse step, we introduce an object-specific sample screening approach to offer sufficient and diverse positive sample candidates for each object. Specifically, we construct a set of Coarse Positive Sample (CPS) candidates for each object, where we consider prior locations from diverse spatial locations and FPN hierarchies as candidates for a specific \textit{gt}. Unlike sampling from a single FPN layer or all FPN layers~\cite{objectbox_2022_eccv,fsaf_2019_cvpr}, we slightly expand the range of candidates to the \textit{gt}'s nearby spatial location and adjacent FPN layers, which warrants relatively diverse and sufficient candidates compared to the single-layer heuristic and narrows down the searching area from all-layer candidates, alleviating tiny object's lack of positive samples candidates.

In this step, we also model the \textit{gt} into a 2-D Gaussian $\mathcal{N}_g(\boldsymbol{\mu}_g, \boldsymbol{\Sigma}_g)$ with the aforementioned method to assist sample selection.
The similarity measurement in constructing the CPS is realized with the Jensen-Shannon Divergence (JSD)~\cite{JS_divergence_TIT_2003} between the anchor and \textit{gt}. JSD inherits the scale invariance property of the Kullback–Leibler Divergence (KLD)~\cite{kld_2021_nips} and can measure the similarity between the \textit{gt} and nearby non-overlapping priors~\cite{kld_2021_nips,rfla_2022_eccv}. Moreover, it overcomes KLD's drawback of asymmetry. However, the closed-form solution of the JSD between Gaussian distributions is unavailable~\cite{gjsd_2020_entropy}. Thus, we use the Generalized Jensen-Shannon Divergence (GJSD)~\cite{gjsd_2020_entropy} which yields a closed-form solution, as the substitute. 
For example, the GJSD between two Gaussian distributions $\mathcal{N}_{p}(\boldsymbol{\mu}_{p}, \boldsymbol{\Sigma}_{p})$ and $\mathcal{N}_{g}(\boldsymbol{\mu}_{g}, \boldsymbol{\Sigma}_{g})$ is defined as follows:
{\small
\begin{equation}
\begin{aligned}
        \mathrm{GJSD}(\mathcal{N}_p, \mathcal{N}_g) &= (1-\alpha) \mathrm{KL}(\mathcal{N}_{\alpha}, \mathcal{N}_{p}) + \alpha \mathrm{KL}(\mathcal{N}_{\alpha}, \mathcal{N}_{g}), \\
                      %&= \frac{1}{2}\left((1-\alpha) \mu_1^{\top} \Sigma_1^{-1} \mu_1+\alpha \mu_2^{\top} \Sigma_2^{-1} \mu_2-\mu_\alpha^{\top} \Sigma_\alpha^{-1} \mu_\alpha+\log \frac{\left|\Sigma_1\right|^{1-\alpha}\left|\Sigma_2\right|^\alpha}{\left|\Sigma_\alpha\right|}\right)
\end{aligned}
\end{equation}
}where $\mathrm{KL}$ denotes the KLD, and $\mathcal{N}_{\alpha}(\boldsymbol{\mu}_\alpha,\boldsymbol{\Sigma}_\alpha)$ is given by:
\begin{equation}
    \boldsymbol{\Sigma}_\alpha=\left(\boldsymbol{\Sigma}_p \boldsymbol{\Sigma}_g\right)_\alpha^{\boldsymbol{\Sigma}}=\left((1-\alpha) \boldsymbol{\Sigma}_p^{-1}+\alpha \boldsymbol{\Sigma}_g^{-1}\right)^{-1},
\end{equation}
and
\begin{equation}
\begin{aligned}
    \boldsymbol{\mu}_\alpha &=\left(\boldsymbol{\mu}_p \boldsymbol{\mu}_g\right)_\alpha^{\boldsymbol{\mu}} \\
    &=\boldsymbol{\Sigma}_{\alpha} \left((1-\alpha) \boldsymbol{\Sigma}_p^{-1}\boldsymbol{\mu}_p+\alpha \boldsymbol{\Sigma}_g^{-1}\boldsymbol{\mu}_g\right),
\end{aligned}
\end{equation}
$\alpha$ is a parameter that controls the weighting of two distributions~\cite{gjsd_2020_entropy} in the similarity measurement. In our case, $\mathcal{N}_p$ and $\mathcal{N}_g$ contribute equally, so $\alpha$ is set to 0.5. 

Ultimately, for each \textit{gt}, we select $K$ priors that hold the top $K$ GJSD scores as the Coarse Positive Samples (CPS) and label the remaining priors as negative samples. This coarse matching serves as the $\mathcal{M}_{s}$ in Equation~\ref{c2f_mapping}. GJSD can effectively measure the similarity between samples across FPN layers with a specific \textit{gt}. Consequently, we extend CPS to include both the object's adjacent region and cross hierarchies by selecting a relatively large number of sample candidates. 

\textbf{Finer posterior matching enhances sample quality.} 
In the fine step, we aim to improve the learning quality without exacerbating the inter-object learning bias. To achieve this, we approximate the instance-wise semantic pattern by representing each object with a Dynamic Gaussian Mixture Model (DGMM). This model serves as the $\mathcal{M}_d$ in Equation~\ref{c2f_mapping} for object-wise sample constraint. Unlike batch-wise or sample-wise evaluations~\cite{dal_2021_aaai,paassignment_2020_eccv,ota_2021_cvpr} which tend to favor larger objects, our approach assesses the relative quality of samples within each object, ensuring consistent positive sample supervision across objects of varying sizes.

First of all, we refine the sample candidates in the CPS according to their predicted scores to fit the object's semantically salient regions. More specifically, we define the Possibility of becoming True predictions ($PT$)~\cite{ota_2021_cvpr} for sample screening, which is a linear combination of the predicted classification score and the location score with the \textit{gt}. We define the $PT$ of the $i^{th}$ sample $D_i$ as: 
\begin{equation}
    PT_i = 0.5( Cls(D_i) + IoU(D_i, gt_i)),
\end{equation}
where $Cls$ is the predicted classification confidence and $IoU$ is the rotated IoU between the predicted location and its corresponding \textit{gt} location. We select candidates with $Q$ highest $PT$ as Medium Positive Sample (MPS) candidates. 

Following this, we define the DGMM using a mixture of \textit{gt's} geometry and MPS distribution to eliminate misaligned samples and obtain the final positive samples for prediction.
Unlike previous works which utilize the center probability map~\cite{CenterMap-Net_2020_TGRS} or the single-Gaussian~\cite{gaussian3d_2022_pami,gghl_2022_tip} for instance representation, our approach represents the instance with a more refined DGMM. This model consists of two components: one centered on the geometry center and the other on the semantic center of the object. Specifically, for a given instance $gt_i$, the geometry center $(cx_i,cy_i)$ serves as the mean vector $\boldsymbol{\mu}_{i,1}$ of the first Gaussian, and the semantic center $(sx_i,sy_i)$, which is deduced by averaging the location of the samples in the MPS, serves as the $\boldsymbol{\mu}_{i,2}$. 
That is to say, we parameterize the instance representation as:
\begin{equation}
    \mathit{DGMM}_i(s|x,y) = \sum_{m=1}^{2}w_{i,m}\sqrt{2\pi|\boldsymbol{\Sigma}_{i,m}|}\mathcal{N}_{i,m}(\boldsymbol{\mu}_{i,m},\boldsymbol{\Sigma}_{i,m}),
\end{equation}
where $w_{i,m}$ is the weight of each Gaussian with a summation of 1, $\boldsymbol{\Sigma}_{i,m}$ equals to the \textit{gt}'s $\boldsymbol{\Sigma}_{g}$.
Under this modeling, each sample in MPS is associated with a DGMM score $\mathit{DGMM}(s|MPS)$. Samples with $\mathit{DGMM}(s|MPS) < e^{-g}$ for any \textit{gt} are assigned negative masks, with $g$ being an adjustable parameter.

%% file: experiments.tex
\section{Experiments}
\label{sec:experiments}
\input{sodaa}
\input{dota2_main_reults}

\subsection{Datasets and Implementations Details}

\textbf{Datasets.} In addition to experiments on the AI-TOD-R, we conduct experiments on seven more datasets covering various tasks to verify the method's broad adaptability. 
These tasks include small oriented object detection (SODA-A~\cite{soda_2023_pami}), oriented object detection with the existence of a large number of tiny objects (DOTA-v1.5~\cite{DOTA2.0_2021_pami}, DOTA-v2~\cite{DOTA2.0_2021_pami}), multi-scale oriented object detection (DOTA-v1~\cite{DOTA_2018_CVPR}, DIOR-R~\cite{diorr_2022_tgrs}), and horizontal object detection (VisDrone~\cite{visdrone2019_2019_iccvw}, MS COCO~\cite{COCO_2014_ECCV}, DOTA-v2 HBB).

For ablation studies and analyses, we choose the large-scale DOTA-v2 {\tt train set} for training and its {\tt val set} for evaluation since DOTA-v2 is the largest dataset for oriented object detection and contains a substantial number of tiny objects. This dataset enables us to simultaneously verify the method's effectiveness on tiny object detection and generic oriented object detection.
For fair comparison with other methods, we use the {\tt trainval sets} of DOTA-v1, DOTA-v1.5, DOTA-v2, and DIOR-R for training and their respective {\tt test sets} for testing, and we use the {\tt train set} and {\tt test set} of SODA-A, the {\tt train sets}, {\tt val sets} of VisDrone2019 and MS COCO for training and evaluation.

\textbf{Implementation details.} We conduct all experiments on a computer equipped with a single NVIDIA RTX 4090 GPU, setting the batch size to 4. The models are built using MMDetection~\cite{mmdetection_2019_arXiv} and MMRotate~\cite{mmrotate_2022_mm} frameworks with PyTorch~\cite{PyTorch_2019_NIPS}. We utilized ImageNet~\cite{ImageNet_2015_IJCV} pre-trained models as the backbone. For training, we employ the Stochastic Gradient Descent (SGD) optimizer with a learning rate of 0.005, momentum of 0.9, and weight decay of 0.0001. Unless otherwise specified, the default backbone is ResNet-50~\cite{ResNet_2016_CVPR} with FPN~\cite{FPN_2017_CVPR}. We use Focal loss~\cite{Focal-Loss_2017_ICCV} for classification and IoU loss~\cite{Unitbox_2016_ACMM} for regression. We only use random flipping for data augmentation across all experiments. 

For experiments on DOTA-v1 and DOTA-v2, we adhere to the official settings of the DOTA-v2 benchmark~\cite{DOTA2.0_2021_pami}. Specifically, we crop images into patches of $1024 \times 1024$ with 200-pixel overlaps and train the models for 12 epochs. For DOTA-v2, we reproduce several state-of-the-art one-stage methods~\cite{FCOS_2019_ICCV,atss_2020_cvpr,orientedrep_2022_cvpr,R3Det_2021_AAAI,kld_2021_nips,dal_2021_aaai,sasm_2022_aaai,s2anet_2021_tgrs} using the same settings. For experiments on other datasets, we follow their default benchmarks for image pre-processings, including setting the input size to $1200 \times 1200$ for SODA-A, $1024 \times 1024$ with 200-pixel overlaps for DOTA-v1.5, $800 \times 800$ for DIOR-R, $1333 \times 800$ for VisDrone and MS COCO. The models are trained for 40 epochs on DOTA-v1.5 and DIOR-R, and for 12 epochs on SODA-A, VisDrone, and MS COCO, following previous works~\cite{beyond_2021_cvpr,orientedrep_2022_cvpr}. The DCFL uses RetinaNet-O as the baseline detector if not specified. Unless otherwise specified, these settings are consistently maintained.

\subsection{Main Results}
\textbf{Tiny/small oriented object detection.} As the main track, we evaluate the performance of DCFL on challenging datasets that are dedicated to tiny (AI-TOD-R) and small (SODA-A) oriented object detection. First of all, results on the AI-TOD-R are shown in Table~\ref{table:supervised}. Without whistles and bells, DCFL can improve both one-stage (\#1 \textit{vs.} \#22) and two-stage object detectors (\#5 \textit{vs.} \#24) by large margins. Notably, when plugging DCFL into the advanced one-stage method: $\rm{S^2A}$-Net, our approach hits a new state-of-the-art performance of 49.6\% $\rm{AP_{0.5}}$, with a remarkable improvement of 16.2\% and significant improvements on very tiny objects. 
Besides, we also evaluate the proposed method on another oriented small object detection benchmark: SODA-A. As a recently proposed dataset, the challenging and large-scale characteristics of the SODA~\cite{soda_2023_pami} attract increasing attention. Results on this benchmark are shown in Table~\ref{tab:sodaa}, where DCFL really shines on this challenging dataset by boosting the RetinaNet-O by 8.1 AP points and boosting the strong baseline: Oriented R-CNN by 2.2 AP points. Moreover, the improvement in terms of $\rm{AP_{0.75}}$ is more pronounced than $\rm{AP_{0.5}}$, indicating that DCFL can more precisely locate the oriented tiny objects. Given that DCFL mainly optimizes the model's training process, the accuracy improvement does not incur additional parameter and computational costs on both datasets, as described in Tables~\ref{table:supervised} and~\ref{tab:sodaa}.

\textbf{Oriented object detection with massive tiny objects.} More generally, evaluating the model's detection performance in datasets with both massive tiny objects and other-sized objects cannot only validate its ability to address tiny objects but also examine its robustness to scale variance. We thus perform experiments on the DOTA-v1.5 and DOTA-v2, which are general-purpose datasets characterized by the existence of a significant number of tiny objects. As shown in Table~\ref{table:dotav2}, our proposed method achieves a state-of-the-art performance of 57.66\% mAP on the challenging DOTA-v2 benchmark with single-scale training and testing. Meanwhile, our model attains 51.57\% mAP on this dataset without bells and whistles, outperforming all tested one-stage oriented object detectors. Besides, results on the DOTA-v1.5 are presented in Table~\ref{tab.dota15}, where DCFL notably improves the baseline and achieves a leading performance among one-stage methods. 

\textbf{Multi-scale oriented object detection.} An investigation of the method's performance on multi-scale oriented object detection datasets can demonstrate its versatility and generality across diverse oriented object detection tasks. 
Therefore, we validate DCFL on the DOTA-v1 and DIOR-R multi-scale oriented object detection datasets, which also include some tiny object classes. The results of these datasets are shown in Tables~\ref{exp.dotav1} and~\ref{diorr}. Beyond tiny object-specific datasets, DCFL also excels in multi-scale scenarios, achieving leading performance among all one-stage methods. Furthermore, the class-wise AP of tiny objects on DOTA-v1 and DIOR-R, listed in Tables~\ref{exp.dotav1} and~\ref{diorr_class}, show particularly significant improvements for tiny-size classes, often with a notable increase of more than 10\%.

\textbf{Horizontal object detection.} The proposed method can also be applied to the generic object detection tasks and enhance their performance, by simply discarding the angle information. We evaluate the model on three different scenarios: drone-captured images (VisDrone), natural images (MS COCO), and aerial images (DOTA-v2 HBB). These datasets, annotated with horizontal bounding boxes, contain a significant number of small objects. Integrating our learning pipeline into the RetinaNet-O baseline results in an improvement of 2-3 points, as shown in Table~\ref{hbb_results}.

In a nutshell, these results demonstrate that our DCFL is not only highly effective for detecting oriented tiny objects (such as small vehicles, ships, and storage tanks), achieving an approximate 10-point improvement over the baseline for these classes. Meanwhile, it excels in general-purpose oriented object detection or horizontal object detection tasks, as evidenced by its performance on tracks like DOTA-v1, DIOR-R, and MS COCO.

\begin{table}[t]
    \centering
    \caption{Comparison with one-stage detectors on the DOTA-v1 OBB Task. All results are based on the MMRotate~\cite{mmrotate_2022_mm} with 12 epochs except for GGHL~\cite{gghl_2022_tip}.~3$\times$ means training for 36 epochs.}
    \resizebox{\linewidth}{!}{
    \begin{tabular}{c|ccccc}  
	\toprule
    Method & CFA~\cite{beyond_2021_cvpr} & RetinaNet-O~\cite{Focal-Loss_2017_ICCV} & $\rm{R^3Det}$~\cite{R3Det_2021_AAAI} & Oriented Rep~\cite{orientedrep_2022_cvpr}  & ATSS-O~\cite{atss_2020_cvpr} \\
    mAP  & 69.63 & 69.79 & 70.18 & 71.94 & 72.29 \\
    \midrule
    Method & KLD~\cite{kld_2021_nips} & $\rm{S^2A}$-Net~\cite{s2anet_2021_tgrs} & GGHL~\cite{gghl_2022_tip} (3$\times$) &  DCFL &  DCFL (3$\times$)\\
    mAP & 72.76 & 73.91  & 73.98 & 74.26 & \textbf{75.35} \\
    \bottomrule
    \end{tabular}}
    \label{exp.dotav1}
\end{table}

\setlength{\tabcolsep}{4pt}
\begin{table}[t]\small
    \centering
    \caption{Main results on the DOTA-v1.5 OBB Task. }
    \resizebox{\linewidth}{!}{
    \begin{tabular}{l|c|ccc|c}  
	\toprule
    Method & Backbone  & SV & Ship & ST  & mAP  \\
    \midrule
    RetinaNet-O~\cite{Focal-Loss_2017_ICCV} & R50 & 44.53 & 73.31 & 59.96 & 59.16 \\
    Faster R-CNN-O~\cite{Mask-R-CNN_2017_ICCV} & R50 & 51.28 & 79.37 & 67.50 & 62.00 \\   
    %MR~\cite{Mask-R-CNN_2017_ICCV} & R50 & 51.31 & 79.75 & 66.07 & 62.67 \\    
    CMR~\cite{Mask-R-CNN_2017_ICCV} & R50 & 51.64 & 79.99 & 67.58 & 63.41 \\
    RoI Transformer~\cite{RoI-Transformer_2019_CVPR} & R50 & 52.05 & 80.72 & 68.26 & 65.03 \\    ReDet~\cite{redet_2021_cvpr} & ReR50 & 52.38 & 80.92 & 68.64 & 66.86 \\
    \midrule
    DCFL & R50 & 56.72  & 80.87  & 75.65  & 67.37 (+8.21) \\
    DCFL & ReR101 & \textbf{57.31} & \textbf{86.60} & \textbf{76.55} & \textbf{70.24} (+11.08) \\
    \bottomrule
    \end{tabular}}
    \label{tab.dota15}
\end{table}

\begin{table}[t]
    \centering
    \caption{Performance comparisons on the DIOR-R dataset.}
    \resizebox{0.95\linewidth}{!}{
    \begin{tabular}{c|cccc}  
	\toprule
    Method & RetinaNet-O~\cite{Focal-Loss_2017_ICCV} & FR-OBB~\cite{Faster-R-CNN_2015_NIPS} & RT~\cite{RoI-Transformer_2019_CVPR} & AOPG~\cite{diorr_2022_tgrs}\\
    mAP  & 57.55 & 59.54  & 63.87 & 64.41\\
    \midrule
    Method & GGHL~\cite{gghl_2022_tip} & Oriented Rep~\cite{orientedrep_2022_cvpr}  & DCFL &  DCFL (ReR101)\\
    mAP  & 66.48 & 66.71 & 66.80 & \textbf{71.03} \\
    \bottomrule
    \end{tabular}}
    \label{diorr}
\end{table}

\begin{table}[t]\small
    \centering
    \caption{Detection results of typical tiny objects on the DIOR-R dataset. VE, BR, and WM denote vehicle, bridge, and wind-mill.}
    \resizebox{0.95\linewidth}{!}{
    \begin{tabular}{l|c|ccc}  
	\toprule
    Method & Backbone & VE & BR & WM \\
    \midrule
    RetinaNet-O~\cite{Focal-Loss_2017_ICCV}  & R50 & 38.0 & 24.0 & 60.2\\
    Oriented Rep~\cite{orientedrep_2022_cvpr}  & R50 & 50.4 & 38.8 & 64.7 \\
    DCFL & R50 & \textbf{50.9} (+12.9) & \textbf{42.1} (+18.1) & \textbf{70.9} (+10.7) \\
    \bottomrule
    \end{tabular}}
    \label{diorr_class}
\end{table}

\begin{table}[t]\footnotesize
    \centering
    \caption{The versatility on generic object detection datasets.}
    \resizebox{\linewidth}{!}{
    \begin{tabular}{l|cc|cc|cc}  
	\toprule
    Dataset & \multicolumn{2}{c|}{VisDrone} & \multicolumn{2}{c|}{MS COCO} & \multicolumn{2}{c}{DOTA-v2 HBB} \\
    \midrule
    Method & RetinaNet~\cite{Focal-Loss_2017_ICCV} & DCFL & RetinaNet & DCFL & FCOS~\cite{rfla_2022_eccv} & DCFL \\
	\midrule
    $\rm{AP}_{0.5}$  & 29.2 & 32.1 & 55.4  & 57.3 &  55.4 & 57.4 \\
	\bottomrule
    \end{tabular}}
    \label{hbb_results}
\end{table}

\begin{figure*}[t]
\centering
\includegraphics[width=0.99\linewidth]{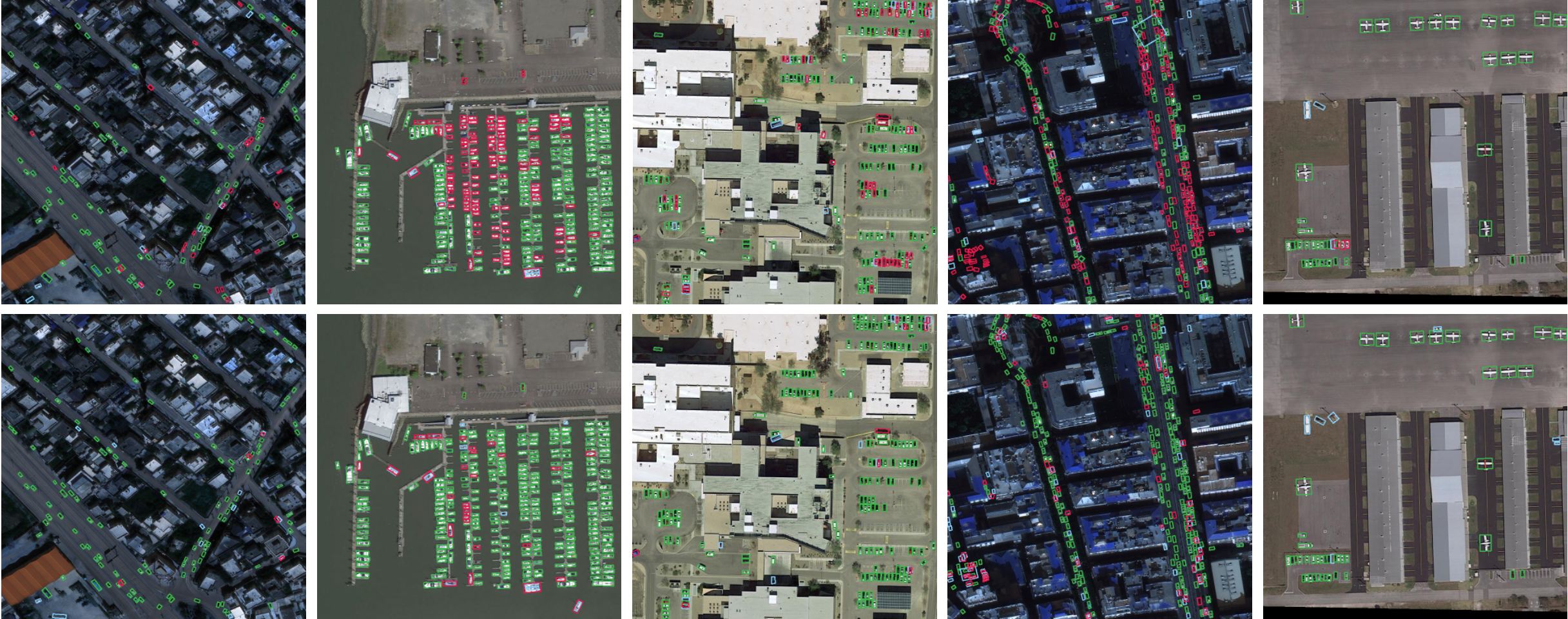}
\caption{Visualization analysis of the predicted results. The first row shows the predicted results by the Oriented R-CNN while the second row shows results from DCFL on the AI-TOD-R dataset. True positive, false negative, and false positive predictions are marked in green, red, and blue, respectively. }
\label{fig:vis_preds}
\end{figure*}

\input{dota2_ablations}

\subsection{Ablations}

\textbf{Effects of individual strategy.} We evaluate the effectiveness of each proposed strategy from our method through a series of ablation experiments. For consistency and fair comparisons, we tile one prior for each feature point in all experiments. As shown in Table~\ref{tab:ablation:individual}, the baseline detector, RetinaNet-O, achieves an mAP of 51.70\%. Gradually integrating the posterior re-ranked MPS and DGMM into the detector, based on the CPS, results in progressive mAP improvements, confirming the effectiveness of each design. It is important to note that CPS cannot be used independently, as its samples are too coarse to serve as the final positive samples. Nevertheless, we compare different methods of constructing the CPS to verify its superiority.   

% need to be revised if we propose to plug it into FCOS
\textbf{Comparisons of different CPS.} The design choice of CPS determines the range of sample candidates when training. In this section, we compare several CPS design paradigms, including limiting the CPS to a specific \textit{gt} within a single layer and utilizing all FPN layers as the CPS, similar to Objectbox~\cite{objectbox_2022_eccv}. We present their performance in Table~\ref{tab:ablation:cps}. For fair comparisons, the number of samples in CPS is fixed at 16, and all other components remain unchanged. In the Single-FPN-layer approach, we group \textit{gt} onto different layers based on the regression range defined in FCOS~\cite{FCOS_2019_ICCV} and assign labels within each layer. In the All-FPN-layer approach, we do not group \textit{gt} onto different layers but instead, discard prior scale information and directly measure the distance between Gaussian \textit{gt} and prior points. As shown in Tab.~\ref{tab:ablation:cps}, neither of these two methods yields the best performance. In contrast, using distribution distances (KLD, GWD, GJSD) to construct the Cross-FPN-layer CPS extends the candidate range to adjacent layers in addition to the main layer. We can also see the GJSD gets the best performance of 59.15\% mAP, mainly resulting from its property of scale-invariance~\cite{kld_2021_nips,gjsd_2020_entropy}, symmetry~\cite{gjsd_2020_entropy}, and ability to measure non-overlapping boxes~\cite{gjsd_2020_entropy} compared to other counterparts.

\textbf{Fixed prior or dynamic prior.} We conduct a detailed set of ablation studies to verify the necessity of introducing the dynamic prior. As shown in Table~\ref{tab:ablation:pcb}, disabling the dynamic prior by fixing the location of samples results in a performance drop. This indicates that the prior should be adjusted accordingly when leveraging the dynamic sampling strategy to better capture the shape of objects.

\textbf{Detailed design in PCB.} The PCB consists of a dilated convolution and a guiding DCN. We slightly enlarge the receptive field using a dilation rate of 3 and then utilize the DCN to generate dynamic priors in a guiding manner. As shown in Table~\ref{tab:ablation:pcb}, the DCN provides an improvement of 0.34 mAP points, and the dilated convolution slightly enhances the mAP. However, applying the DCN~\cite{dcnv2_2019_cvpr} to the single regression branch slightly deteriorates accuracy (denoted as \textit{Separate} in Table~\ref{tab:ablation:pcb}), likely due to mismatch issues between the two branches. To address this, we use the offsets from the regression head to guide the offsets for the classification head, resulting in better alignment (denoted as \textit{Guiding}).

\textbf{Effects of parameters.} The three introduced parameters are robust within a certain range. As shown in Table~\ref{tab:ablation:kq}, the combination of $K=16$ and $Q=12$ yields the best performance. In Table~\ref{tab:ablation:g}, we verify the threshold $e^{-g}$ in the DGMM and find that setting $w_{i,1}$ to 0.7 and a threshold of $g=0.8$ results in the highest mAP. Although making the CPS/MPS/DGMM coarser and stricter can weaken performance, the mAP only fluctuates slightly. This indicates that the coarse-to-fine assignment method ensures robustness in parameter selection, as multiple parameters can mitigate the effects of any single under-tuned parameter.

\begin{figure}[t]
\centering
\includegraphics[width=\linewidth]{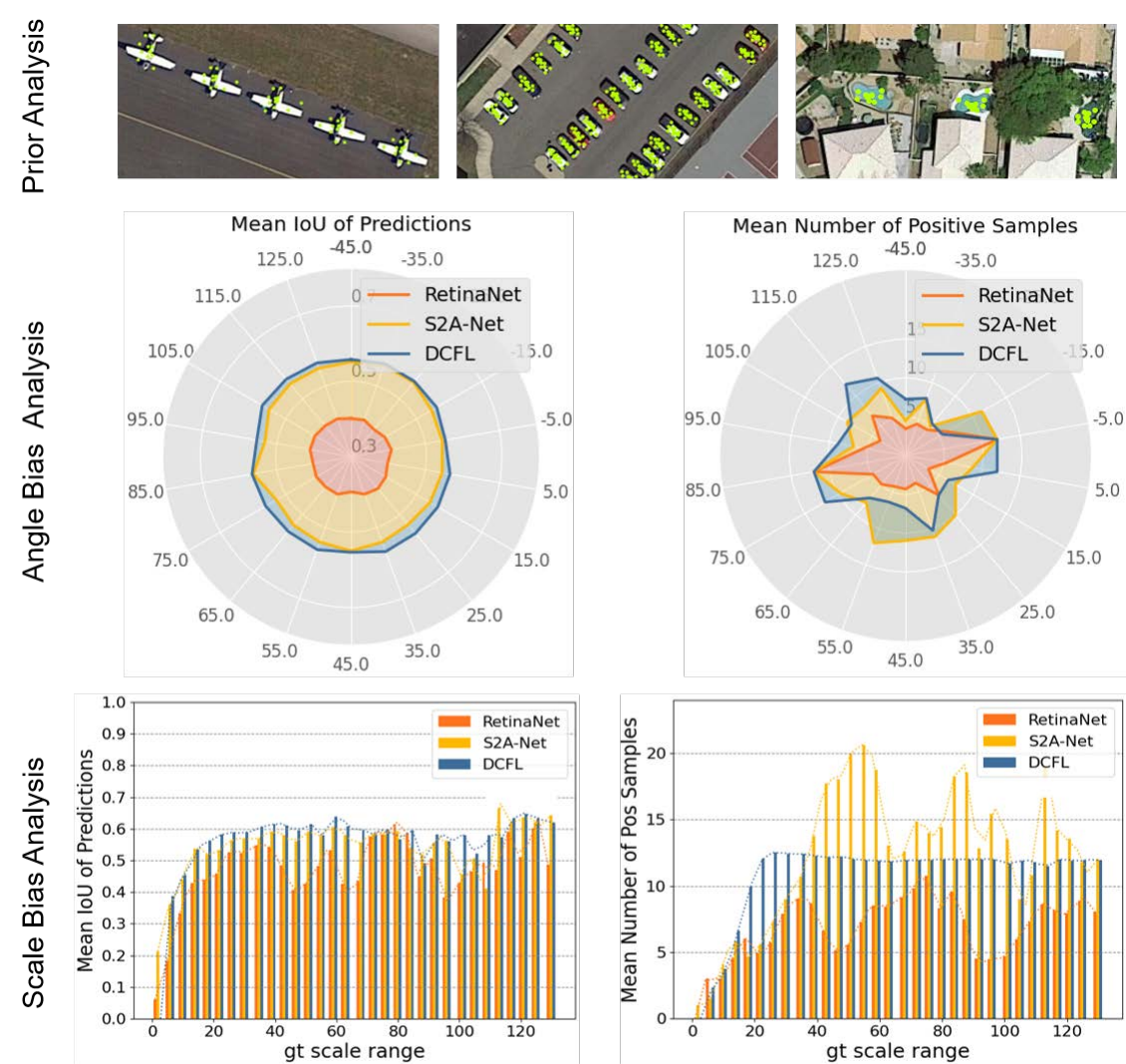}
\caption{Analysis of the learning bias across different methods. The first and second columns investigate quality and quantity imbalances,  respectively. Results are sampled from the model's last training epoch.}
\label{analysis_bias}
\end{figure}

\subsection{Analysis}
\textbf{Visual Analysis.} We visualize DCFL's predictions and dynamic prior positions to better show models' capability on addressing oriented tiny objects in Figures~\ref{fig:vis_preds} and~\ref{analysis_bias}, respectively. In Figure~\ref{fig:vis_preds}, By separating the model's predictions into true positive, false negative, and false positive predictions based on the \textit{gt} with different colors, we can easily find that DCFL significantly suppresses false negative predictions (\textit{i.e.}, missing detection) for tiny objects. This improvement can be largely attributed to the sufficient and unbiased sample learning of different-sized objects resulting from the coarse-to-fine sample selection scheme.
Besides, from Figure~\ref{analysis_bias} (Upper), we can find that the prior setting in DCFL can better match the oriented tiny objects' discriminative areas. This further verifies that by adaptively adjusting prior positions according to the object's region of interest, the prior bias in previous static prior designs can be mitigated. 

\textbf{How does DCFL achieve unbiased learning?}
To better understand the working mechanism of DCFL, we delve into its training process by statistically investigating its sample assignment. Specifically, we calculate the quantity and quality of positive samples assigned to ground truth (\textit{gt}) bounding boxes within various angle and scale intervals. This analysis reveals two types of imbalance issues (quantity and quality) in baseline methods: (1) The number of positive samples assigned to each object varies periodically with respect to its angle and scale, with objects whose shapes (scale, angle) differ from predefined anchors receiving much fewer positive samples. (2) The predicted IoU fluctuates periodically with respect to the \textit{gt}'s scale while remaining invariant with respect to the \textit{gt}'s angle.
In contrast, DCFL effectively addresses these learning biases: (1) It compensates by assigning more positive samples to previously outlier angles and scales. (2) It improves and balances the quality of samples (predicted IoU) across all angles and scales. These results demonstrate the desired behavior of dynamic coarse-to-fine learning.

\section{Discussions}
The precise detection of arbitrary-oriented tiny objects is a fundamental step towards more generic pattern recognition in numerous specialized scenarios. Meanwhile, the state-of-the-art object detectors significantly degrade when detecting these objects. Moreover, there is still a lack of task-specific datasets and benchmarks dedicated to corresponding research. This motivates us to address this intricate but inevitable challenge. To this end, we establish a task-specific dataset, benchmark, and design a new method that realizes unbiased learning for objects of different scales and orientations. 

Nevertheless, some challenges remain. First, the detection of oriented tiny objects is a widespread issue across various scenarios (\textit{e.g.}, autonomous driving, medical imaging, and defect detection) and diverse modalities (\textit{e.g.}, SAR, thermal, and X-ray data). This work, however, primarily focuses on aerial scenes in high-resolution optical data. By focusing on the typical scenario of aerial imagery where oriented tiny objects frequently appear, we aim to establish a solid foundation and open the possibility for understanding these challenging objects in a broader range of scenarios and modalities. Future research could also explore incorporating complementary information from different modalities or leveraging temporal data to enhance the detection of oriented tiny objects, further expanding and fulfilling practical applications.
Second, the methodology part in this paper is performed under the closed-set setting, which requires full object annotations from the training set. However, the object annotations for tiny objects with oriented information are scarce and their acquisition process is difficult, especially when it comes to scenarios in an open-world assumption. Meanwhile, experimental results have shown that label-efficient methods show very competitive performance compared to fully-supervised methods on oriented tiny object detection. Thus, it is worth further exploring the simplification of annotation requirements and the enhancement of tiny object detection performance with limited annotations.
Third, foundation models are becoming a hot topic that facilitates various research directions while this work did not discuss or improve relevant works. How foundation models perform and how to pre-train or adapt them on oriented tiny objects are also questions worth exploring in the future.

%% file: sodaa.tex
\setlength{\tabcolsep}{3pt}
\begin{table*}[t]
	\centering
	\caption{Results on SODA-A test-set. All the models are trained on SODA-A train-set with a ResNet-50 as the backbone. Schedule denotes the training epochs, where '1$\times$' refers to 12 epochs.}
	\resizebox{\linewidth}{!}{
		\begin{tabular}{l|c|c|ccc|cccc|c|c}
			\hline
			Method & Publication & Schedule & $\rm{AP}$  & $\rm{AP_{0.5}}$  & $\rm{AP_{0.75}}$  & $\rm{AP_{eS}}$  & $\rm{AP_{rS}}$  & $\rm{AP_{gS}}$  & $\rm{AP_{N}}$  & $\#$Params. & FLOPs \\
			\hline
			Faster RCNN-O \cite{Faster-R-CNN_2015_NIPS} & TPAMI 2017 & 1$\times$    & 32.5 & 70.1  & 24.3  & 11.9  & 27.3  & 42.2 & 34.4 & 41.1M & 292.25G \\
			RetinaNet-O \cite{Focal-Loss_2017_ICCV} & TPAMI 2020 & 1$\times$    & 26.8  & 63.4  & 16.2  & 9.1  & 22.0  & 35.4  & 28.2 & 36.2M & 221.90G \\
			RoI Transformer \cite{RoI-Transformer_2019_CVPR} & CVPR 2019 & 1$\times$    & 36.0  & 73.0  & 30.1  & 13.5  & 30.3  & 46.1 & 39.5  & 55.1M & 306.20G \\
			Gliding Vertex \cite{Gliding_Vertex_2020_PAMI} & TPAMI 2021 & 1$\times$    & 31.7 & 70.8 & 22.6 & 11.7 & 27.0 & 41.1 & 33.8 & 41.1M & 292.25G \\
			Oriented RCNN \cite{orientedrcnn_2021_iccv} & ICCV 2021 & 1$\times$    & 34.4  & 70.7  & 28.6  & 12.5  & 28.6  & 44.5 & 36.7 & 41.1M & 292.44G \\
			S$^2$A-Net \cite{s2anet_2021_tgrs} & TGRS 2022 & 1$\times$    & 28.3  & 69.6  & 13.1  & 10.2  & 22.8  & 35.8 & 29.5 & 38.6M & 277.72G \\
			DODet \cite{dodet_tgrs_2022} & TGRS 2022 & 1$\times$  & 31.6 & 68.1 & 23.4 & 11.3 & 26.3 & 41.0 & 33.5 & 69.3M & 555.49G \\
			Oriented RepPoints \cite{orientedrep_2022_cvpr} & CVPR 2022 & 1$\times$  & 26.3  & 58.8  & 19.0  & 9.4  & 22.6  & 32.4  & 28.5 & 55.7M & 274.07G \\
			DHRec \cite{DHRec_2023_tpami} & TPAMI 2022 & 1$\times$  & 30.1 & 68.8 & 19.8 & 10.6 & 24.6 & 40.3 & 34.6 & 32.0M & 792.76G \\
   			CFINet~\cite{cfinet_iccv_2023} & ICCV 2023 & 1$\times$  & 34.4 & 73.1 & 26.1 & 13.5 & 29.3 & 44.0 & 35.9 & 44.0M & 312.60G \\
			\hline
   			DCFL (RetinaNet-O) & Ours & 1$\times$  & 34.9 (+8.1) & \textbf{73.2} (+9.8) & 27.8 (+11.6) & \textbf{14.2} & 29.8 & 43.7 & 38.0 & 36.1M & 221.90G \\
   			DCFL (Oriented R-CNN) & Ours & 1$\times$  & \textbf{36.6} (+2.2) & 72.6 (+1.9) & \textbf{32.4} (+3.8) & 13.9 & \textbf{30.3} & \textbf{47.4} & \textbf{41.2} & 41.1M & 292.44G \\
            \hline
		\end{tabular}%
		\label{tab:sodaa}%
	}
\end{table*}%

%% file: dota2_main_reults.tex
\setlength{\tabcolsep}{1.5pt}
\begin{table*}[ht]
\caption{Main results on the DOTA-v2 OBB Task. We follow the official class abbreviations as the DOTA-v2.0 benchmark~\cite{DOTA2.0_2021_pami}. Dp denotes Deformable RoI Pooling~\cite{DCN_2017_CVPR}. \dag~denotes training for 40 epochs. Note that this paper~\cite{kld_2021_nips} reports 50.90\% mAP for R$^3$Det~w/~KLD under 20 epochs, the ReR101 backbone is proposed by the ReDet~\cite{redet_2021_cvpr}. The results in \textbf{bold} and \underline{underline} denote the best and second-best performance of each column.}
\footnotesize
\begin{center}
\resizebox{\linewidth}{!}{
\begin{tabular}{l|c|cccccccccccccccccc|c}  
	\toprule
	Method  & Backbone & Plane & BD & Bridge & GTF & SV & LV & Ship & TC & BC & ST & SBF & RA & Harbor & SP & HC & CC & Air & Heli & mAP  \\
	\midrule
        \textit{multi-stage:}      & & 	&  &  &  &  &  &  &   & & 	&  &  &  &  &  & & & \\
    Faster R-CNN-O~\cite{Faster-R-CNN_2015_NIPS} & R50 & 71.61 & 47.20 & 39.28 & 58.70 & 35.55 & 48.88 & 51.51 & 78.97 & 58.36 & 58.55 & 36.11 & 51.73 & 43.57 & 55.33 & 57.07 & 3.51 & 52.94 & 2.79 & 47.31 \\
    Faster R-CNN-O w/ Dp &  R50 & 71.55 & 49.74 & 40.34 & 60.40 & 40.74 &50.67 &56.58 &79.03 &58.22& 58.24 &34.73 & 51.95 & 44.33 & 55.10 & 53.14 & 7.21 & 59.53 & 6.38 & 48.77 \\
    Mask R-CNN~\cite{Mask-R-CNN_2017_ICCV}  & R50 & 76.20 & 49.91 & 41.61 & 60.00 & 41.08 & 50.77 & 56.24 & 78.01 & 55.85 & 57.48 & 36.62 & 51.67 & 47.39 & 55.79 & 59.06 & 3.64 & 60.26 & 8.95 & 49.47   \\
    %CMR & R50 &  77.01 & 47.54 & 41.79 & 58.02 & 41.58 & 51.74 & 57.86 & 78.2 & 56.75 & 58.5 & 37.89 & 51.23 & 49.38 & 55.98 & 54.59 & 12.31 & 67.33 & 3.01 & 50.04 \\
    HTC*~\cite{HTC_2019_CVPR} & R50 & 77.69 & 47.25 & 41.15 & 60.71 & 41.77 & 52.79 & 58.87 & 78.74 & 55.22 & 58.49 & 38.57 & 52.48 & 49.58 & 56.18 & 54.09 & 4.20 & 66.38 & 11.92 & 50.34 \\
    RoI Transformer~\cite{RoI-Transformer_2019_CVPR} & R50 &  71.81 & 48.39 & 45.88 & 64.02 & 42.09 & 54.39 & 59.92 & \textbf{\textbf{82.70}} & \underline{{63.29}} & 58.71 & 41.04 & 52.82 & 53.32 & 56.18 & 57.94 & 25.71 & 63.72 & 8.70 & 52.81 \\
    Oriented R-CNN~\cite{orientedrcnn_2021_iccv} & R50 & 77.95 & 50.29 & \underline{{46.73}} & \underline{{65.24}} & 42.61 & 54.56 & \textbf{\textbf{60.02}} & 79.08 & 61.69 & 59.42 & 42.26 & \textbf{\textbf{56.89}} & 51.11 & 56.16 & \underline{{59.33}} & 25.81 & 60.67 & 9.17 & 53.28\\
    \midrule
      \textit{one-stage:}   & & 	&  &  &  &  &  &  &   & & 	&  &  &  &  &  & & &    \\
    DAL~\cite{dal_2021_aaai} & R50 & 71.23 & 38.36 & 38.60 & 45.24 & 35.42 & 43.75 & 56.04 & 70.84 & 50.87 & 56.63 & 20.28 & 46.53 & 33.49 & 47.29 & 12.15 & 0.81 & 25.77 & 0.00 & 38.52 \\
    SASM~\cite{sasm_2022_aaai} & R50 & 70.30 & 40.62 & 37.01 & 59.03 & 40.21 & 45.46 & 44.60 & 78.58 & 49.34 & 60.73 & 29.89 & 46.57 & 42.95 & 48.31 & 28.13 & 1.82 & \underline{{76.37}} & 0.74 & 44.53 \\
	RetinaNet-O~\cite{Focal-Loss_2017_ICCV}  & R50 & 70.63 & 47.26 & 39.12 & 55.02 & 38.10 & 40.52 & 47.16 & 77.74 & 56.86 & 52.12 & 37.22 & 51.75 & 44.15 & 53.19 & 51.06 & 6.58 & 64.28 & 7.45 & 46.68    \\
    $\rm{R^3Det}$~w/~KLD~\cite{kld_2021_nips} & R50 & 75.44 & 50.95 & 41.16 & 61.61 & 41.11 & 45.76 & 49.65 & 78.52 & 54.97 & 60.79 & 42.07 & 53.20 & 43.08 & 49.55 & 34.09 & \textbf{\textbf{36.26}} & 68.65 & 0.06 & 47.26 \\
    FCOS-O~\cite{FCOS_2022_TPAMI} & R50 & 74.84 & 47.53 & 40.83 & 57.41 & 43.89 & 47.72 & 55.66 & 78.61 & 57.86 & 63.00 & 38.02 & 52.38 & 41.91 & 53.24 & 40.22 & 7.15 & 65.51 & 7.42 & 48.51 \\
    Oriented Reppoints~\cite{orientedrep_2022_cvpr} & R50 & 73.02 & 46.68 & 42.37 & 63.05 & 47.06 & 50.28 & 58.64 & 78.84 & 57.12 & \textbf{\textbf{66.77}} & 35.21 & 50.76 & 48.77 & 51.62 & 34.23 & 6.17 & 64.66 & 5.87 & 48.95\\
    ATSS-O~\cite{atss_2020_cvpr} & R50 & 77.46 & 49.55 & 42.12 & 62.61 & 45.15 & 48.40 & 51.70 & 78.43 & 59.33 & 62.65 & 39.18 & 52.43 & 42.92 & 53.98 & 42.70 & 5.91 & 67.09 & 10.68 & 49.57 \\
	$\rm{S^2A}$-Net~\cite{s2anet_2021_tgrs}  & R50 & 77.84 & 51.31 & 43.72 & 62.59 & 47.51 & 50.58 & 57.86 & 80.73 & 59.11 & 65.32 & 36.43 & 52.60 & 45.36 & 52.46 & 40.12 & 0.00 & 62.81 & 11.11 & 49.86    \\
    %$\rm{R^3Det + KLD}$ & R50 & - & - & -& - & - & - & - & - & - & - & - & -& - & - & - & - & - & - & 50.90 \\
	\midrule
    \textit{ours:}   & & 	&  &  &  &  &  &  &   & & 	&  &  &  &  &  & & &    \\
     DCFL (Retinanet-O)  & R50 & 75.71 & 49.40 & 44.69 & 63.23 & 46.48 & 51.55 & 55.50 & 79.30 & 59.96 & 65.39 & 41.86 & 54.42 & 47.03 & 55.72 & 50.49 & 11.75 & 69.01 & 7.75 & 51.57 \\
    DCFL ($\rm{S^2A}$-Net)  & R50 & 74.79 & \underline{{53.25}} & 45.81 & \textbf{\textbf{65.46}} & 46.49 & 53.23 & 58.10 & \underline{{81.51}} & 60.13 & \underline{{66.42}} & 43.24 & 55.09 & 50.52 & 55.58 & 54.53 & 5.23 & 68.73 & 13.06 & 52.84   \\
    DCFL (Oriented R-CNN)  & R50 & 77.59 & 52.46 & 45.98 & 61.73 & 49.77 & 54.32 & 60.55 & 79.27 & 61.76 & 68.17 & 43.41 & 56.59 & 52.41 & 56.68 & 55.32 & 27.42 & 63.50 & 12.64 & 54.42\\
    DCFL (Retinanet-O)\dag   & R50 & \underline{{78.30}} & 53.03 & 44.24 & 60.17 & \underline{{48.56}} & \textbf{\textbf{55.42}} & 58.66 & 78.29 & 60.89 & 65.93 & \underline{{43.54}} & 55.82 & \underline{{53.33}} & \textbf{\textbf{60.00}} & 54.76 & 30.90 & 74.01 & \textbf{\textbf{15.60}} & \underline{{55.08}} \\
    DCFL (Retinanet-O)\dag   & ReR101 & \textbf{\textbf{79.49}} & \textbf{\textbf{55.97}} & \textbf{\textbf{50.15}} & 61.59 & \textbf{\textbf{49.00}} & \underline{{55.33}} & \underline{{59.31}} & 81.18 & \textbf{\textbf{66.52}} & 60.06 & \textbf{\textbf{52.87}} & \underline{{56.71}} & \textbf{\textbf{57.83}} & \underline{{58.13}} & \textbf{\textbf{60.35}} & \underline{{35.66}} & \textbf{\textbf{78.65}} & \underline{{13.03}} & \textbf{\textbf{57.66}} \\
	\bottomrule
	\end{tabular}}
\label{table:dotav2}
\end{center}
\end{table*}
\setlength{\tabcolsep}{1pt}

%% file: dota2_ablations.tex
\begin{table*}[t]
% subfloat a - BackBone Architecture
\centering
\caption{\textbf{Ablations}. We train on DOTA-v2 train set, test on its val set, and report mAP under IoU threshold 0.5.}
\subfloat[\textbf{Individual effectiveness.} CPS, MPS, and DGMM denote Coarse, Medium Sample Candidates and Dynamic Gaussian Mixture Model.\label{tab:ablation:individual}]{
\tablestyle{4pt}{1.03}
\begin{tabular}{c|ccc|c}  
	\toprule
    Method & CPS & MPS & DGMM & mAP \\
    \midrule
    baseline~\cite{Focal-Loss_2017_ICCV}  &    &  &  &  51.70\\    
    \midrule
      & \checkmark  &  \checkmark  &  & 53.41 \\
     DCFL &  \checkmark  &  &  \checkmark  & 57.20 \\
      &  \checkmark  &  \checkmark  &  \checkmark  & \textbf{59.15} \\
    \bottomrule
    \end{tabular}}\hspace{3mm}
% subfloat b - Multinomial vs Independent Masks
\subfloat[\textbf{Comparsions of different CPS.} The FPN layer number varies for different strategies of getting the CPS.\label{tab:ablation:cps}]{
\tablestyle{4.8pt}{0.95}\begin{tabular}{c|c|c}  
	\toprule
    Strategy & Measurement & mAP  \\
    \midrule
     %base: MaxIoU  & IoU & 51.70 \\
     All-FPN-layer &  Gaussian &  50.12 \\
     Single-FPN-layer &  Gaussian & 56.72 \\
     %Single-FPN-layer &  Center &  \\
     %All-FPN-layer &  Center &  \\
     Cross-FPN-layer & KLD~\cite{kld_2021_nips} &  57.82 \\
     Cross-FPN-layer & GWD~\cite{gwd_2021_icml} &  58.55\\
     Cross-FPN-layer & GJSD &  \textbf{59.15}\\
    \bottomrule
    \end{tabular}}\hspace{3mm}
% subfloat c - RoIAlign (ResNet-50-C4)
\subfloat[\textbf{Effects of designs in the PCB.}\label{tab:ablation:pcb} DP: the dynamic prior. Guiding: \textit{reg} guides \textit{cls} branch.]{
\tablestyle{2.2pt}{0.95}\begin{tabular}{ccc|c}  
	\toprule
    DCN & Dilated Conv & DP & mAP \\
    \midrule
     & & & 58.07 \\
    \checkmark & & & 58.41\\
    \checkmark & \checkmark & & 58.65\\
    \textit{Separate} & \checkmark & \checkmark & 58.71 \\
    \textit{Guiding} & \checkmark & \checkmark & \textbf{59.15} \\
    \bottomrule
    \end{tabular}}\\
% subfloat d - Effects of parameters 
\subfloat[\textbf{Effects of parameters $K$ and $Q$.}\label{tab:ablation:kq}]{
\tablestyle{4.5pt}{0.55}\begin{tabular}{c|cccc|cccc}  
	\toprule
    $K$ & \multicolumn{4}{|c|}{24} & \multicolumn{4}{c}{20}  \\
    \midrule
    $Q$  &  20 & 16 & 12 & 8 &  16 & 12 & 10 & 8 \\
    \midrule
    mAP  &  58.31 & 58.11 & 58.95 & 59.06 &  58.66 & 58.71 & 58.92 & 58.28 \\
    \midrule
    \midrule
    $K$ & \multicolumn{4}{|c|}{16} & \multicolumn{4}{c}{12}   \\
    \midrule
    $Q$  &  12 & 10 & 8 & 6 &  10 & 8 & 6 & 4 \\
    \midrule
    mAP  &  \textbf{59.15} & 58.57 & 58.97 & 57.84 &  58.79 & 58.25 & 57.01 & 57.37\\
    \bottomrule
    \end{tabular}}\hspace{3mm}
% subfloat e - mask representation
\subfloat[\textbf{Effects of parameter $g$}.\label{tab:ablation:g}]{
\tablestyle{12pt}{1.45}\begin{tabular}{c|cc}  
    	\toprule    
        $g$ & 1.2 & 1.0 \\
        \midrule
        mAP & 57.91 &  58.20  \\
        \midrule
        \midrule
        $g$ & 0.8 & 0.4\\
        \midrule
        mAP & \textbf{59.15} & 58.95 \\
        \bottomrule
        \end{tabular}}
% main caption
\label{tab:ablations}
\end{table*}